\documentclass[journal,10pt]{IEEEtran}
\usepackage{hyperref}

\ifCLASSINFOpdf
\else
\fi
\usepackage{color}

\usepackage[utf8]{inputenc} 
\usepackage[T1]{fontenc}    
\usepackage{hyperref}       
\usepackage{url}            
\usepackage{booktabs}       
\usepackage{amsfonts}       
\usepackage{nicefrac}       
\usepackage{microtype}      

\definecolor{Sijia_color}{rgb}{0.858, 0.188, 0.478}

\newcommand{\PY}[1]{\textcolor{green}{PY: #1}}

\newcommand{\BK}[1]{\textcolor{red}{BK: #1}}

\usepackage{floatrow}
\newfloatcommand{capbtabbox}{table}[][\FBwidth]

\usepackage{blindtext}
\usepackage{wrapfig}
\usepackage{array}

\usepackage[small,bf]{caption}
\usepackage{mathtools}
\usepackage{cite}

\usepackage{color}
\usepackage{lipsum}
\usepackage[textsize=scriptsize]{todonotes}

\usepackage{booktabs}       
\usepackage{amsfonts}       
\usepackage{nicefrac}       
\usepackage{microtype}      
\usepackage{makecell}
\usepackage{adjustbox}
\usepackage[para,online,flushleft]{threeparttable}

\usepackage{bbding}
\usepackage{pifont}
\usepackage{wasysym}
\usepackage{amssymb}

\usepackage{multirow}
\usepackage{grffile}
\usepackage{url}
\usepackage{pifont}
\usepackage{comment}

\usepackage{algorithm}
\usepackage{algorithmic}
\usepackage{setspace}
\AtBeginDocument{%
  \addtolength\abovedisplayskip{-0.4\baselineskip}%
  \addtolength\belowdisplayskip{-0.4\baselineskip}%
  \addtolength\abovedisplayshortskip{-0.4\baselineskip}%
  \addtolength\belowdisplayshortskip{-0.4\baselineskip}%
}

\DeclareMathOperator{\diag}{diag}

\DeclareMathOperator*{\minimize}{\text{minimize}}
\DeclareMathOperator*{\maximize}{\text{maximize}}
\newcommand{\argmin}{\operatornamewithlimits{arg\,min}}

\DeclarePairedDelimiterX{\inp}[2]{\langle}{\rangle}{#1, #2}

\DeclareMathOperator*{\st}{\text{subject to}}
\DeclareMathAlphabet\mathbfcal{OMS}{cmsy}{b}{n}
\newcommand{\Def}[0]{\mathrel{\mathop:}=}

\usepackage{xcolor}

\usepackage{romannum}

\begin{document}
%
\title{A Primer on Zeroth-Order Optimization in Signal Processing and Machine Learning}
%
%
%

\author{
Sijia~Liu,~\IEEEmembership{Member,~IEEE,}
        Pin-Yu~Chen,~\IEEEmembership{Member,~IEEE,}
        Bhavya~Kailkhura,~\IEEEmembership{Member,~IEEE,}
     Gaoyuan~Zhang,
        Alfred~Hero,~\IEEEmembership{Fellow,~IEEE,}
        and~Pramod~K.~Varshney,~\IEEEmembership{Life Fellow,~IEEE}
\thanks{S. Liu, P.-Y. Chen and G. Zhang are with the MIT-IBM Watson AI Lab, IBM Research, USA. E-mail: \{sijia.liu, pin-yu.chen, gaoyuan.zhang\}@ibm.com}
\thanks{B. Kailkhura is with Lawrence Livermore National Laboratory, USA. E-mail: kailkhura1@llnl.gov}
\thanks{Alfred O. Hero III is with   University of Michigan, Ann Arbor, USA. E-mail: hero@umich.edu}
\thanks{P. K. Varshney is with Syracuse University, USA. E-mail: varshney@syr.edu}
}

\maketitle

\begin{abstract}
Zeroth-order (ZO) optimization is a subset of gradient-free optimization that emerges in many signal processing and machine learning applications. It is used for solving optimization problems similarly to gradient-based methods. However, it does not require the gradient, using only function evaluations. Specifically, ZO optimization iteratively performs three major steps: gradient estimation, descent direction computation, and solution update. In this paper, we provide a comprehensive review of ZO optimization, with an emphasis on showing the underlying intuition, optimization principles  and  recent advances in convergence analysis. Moreover, we  demonstrate  promising applications of ZO optimization, such as evaluating robustness and generating explanations from black-box deep learning models, and efficient online sensor management. 
\end{abstract}

\begin{IEEEkeywords}
Zeroth-order (ZO) optimization, nonconvex optimization, gradient estimation, black-box adversarial attacks,  machine learning, deep learning
\end{IEEEkeywords}

%
\IEEEpeerreviewmaketitle

\section{Introduction}

Many signal processing,   machine learning (ML) and deep learning (DL) applications
involve tackling complex optimization problems that are difficult to solve analytically. Often the objective function itself may not be in analytical closed form, only permitting function evaluations but not gradient evaluations.
Optimization corresponding to  these types of problems falls  into
the category of 
{zeroth-order (ZO) optimization} with respect to black-box models, 
  where explicit expressions of the gradients are difficult {to compute} or infeasible to obtain.
 ZO optimization methods are gradient-free counterparts of    first-order (FO)  optimization methods.
 They approximate
  the    full gradients or stochastic gradients  through function value based gradient estimates. Interest in  ZO optimization has grown rapidly in the past few years since 
  the concept of gradient estimation by finite difference approximations was proposed in the $1950$s and $1980$s \cite{kiefer1952stochastic,spall1987stochastic}.
  

\begin{figure}[!t]
\centering
\includegraphics[width=0.95\columnwidth]{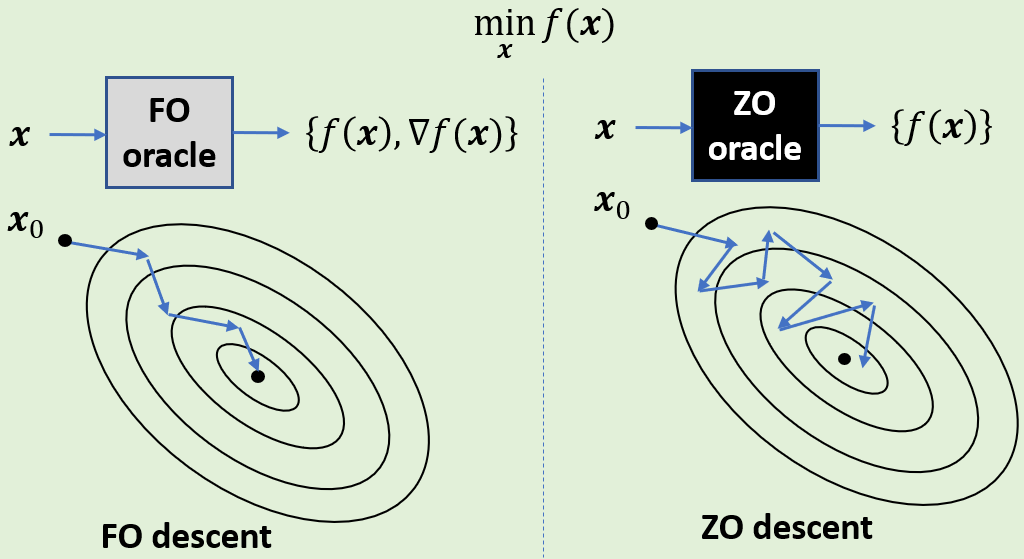}
		\caption{An illustration of FO optimization (left plot) versus ZO optimization (right plot). Here the former solves the optimization problem $\min_{\mathbf x} ~ f(\mathbf x)$ with the white-box objective function $f$, and the latter solves the problem when $f$ is a black-box function. Typically, ZO optimization has a slower convergence speed than FO optimization. 
        }
		\label{fig: ZO_FO}
\end{figure}

It is worth noting that  derivative-free methods for black-box optimization  had been studied by
the optimization community long before they had   impact
on signal processing and ML/DL. 
 Traditional derivative-free optimization (DFO)  methods can 
be  classified into two categories: direct search-based methods (DSMs) and model-based methods (MBMs) \cite{nocedal2006numerical,conn2009introduction,rios2013derivative,Larson_2019}. DSMs include the     
Nelder-Mead simplex method \cite{nelder1965simplex},
the coordinate search method \cite{fermi1942}, and  the  pattern search method \cite{torczon1991convergence}, to name a few. MBMs contain  model-based descent methods \cite{bortz1998simplex} and trust region methods \cite{conn2000trust}.   
Evolutionary optimization is another class of   generic population-based meta heuristic DFO algorithms, and includes particle swarm optimization methods \cite{vaz2009pswarm} and genetic algorithms \cite{goldberg1988genetic}.
Some Bayesian optimization (BO) methods  
\cite{shahriari2016taking} 
tackle black-box optimization problems by
  modelling the objective function as a Gaussian process (GP) that is learned from the history of function evaluations. 
  However, 
learning an  accurate GP model is computationally intensive. 

Conventional DFO methods have two main shortcomings. First, they are often difficult to scale to  large-size problems.  For example,  the off-the-shelf DFO solver \textsf{COBYLA} \cite{powell1994direct} only supports problems with a maximum  of $2^{16}$ variables    (SciPy  Python library \cite{scipy2001}), which is  smaller than the size of a single ImageNet image \cite{deng2009imagenet}. Second, DFO methods lack a convergence rate analysis and they may require  a significant amount of effort to be customized to the particular applications. 
ZO optimization has
  three main  advantages over DFO: a) ease of implementation with only small  modification of  commonly-used gradient-based algorithms, 
  b) computationally efficient approximations to  derivatives when they are difficult   to compute, 
  and c) comparable convergence rates   to FO algorithms  
\cite{ghadimi2013stochastic,nesterov2015random,flaxman2005online,duchi2015optimal}. An illustrative example of ZO optimization versus FO optimization is shown in Figure\,\ref{fig: ZO_FO}.

  ZO  optimization  has attracted increasing attention due to its success in   solving emerging signal processing and ML/DL
problems. 
First, 
ZO optimization serves as a powerful {and practical} tool for evaluating 
adversarial robustness 
of  ML/DL systems \cite{goodfellow2015explaining}. We note that the research in adversarial robustness   
   is receiving increased attention in recent years.
 ZO based methods for exploring vulnerability of DL to {black-box adversarial
attacks} are able to reveal the most susceptible features. Such ZO methods can be  
as effective as state-of-the-art white-box attacks, despite
  only having access to the inputs and outputs of the targeted deep neural networks (DNNs) \cite{chen2017zoo,ilyas2018blackbox}. 
Moreover,  ZO optimization can generate explanations and  provide interpretations of prediction results in  a gradient-free and model-agnostic  manner \cite{dhurandhar2019model}. Furthermore,
ZO optimization  can also be used to solve automated ML problems, e.g., 
automated backpropagation in DL,
where the gradients with respect to ML pipeline configuration parameters are intractable \cite{liu2019admm}.
ZO optimization is also applicable to   ML  applications where the full gradient must be kept private \cite{liu2017zeroth}. 
In addition, 
 ZO optimization   provides {computationally-efficient alternatives for second-order optimization} such as  {robust   training} by  curvature regularization \cite{moosavi2019robustness},  
 meta-learning \cite{Song2020ES-MAML}, transfer learning \cite{Tsai_BAR}, 
 and
 {online network  management} \cite{liu2017zeroth}.
 In this paper, we provide a 
  comprehensive review of recent development in ZO optimization for signal processing and ML. 
In Sections \ref{sec: grad_est} and  \ref{sec: ZO_algs},
 we  review various types of ZO gradient estimators as well as ZO algorithms.
 Section\,\ref{sec: adv} presents a promising connection between ZO optimization and adversarial ML.  
 Section\,\ref{sec: ZO_sensor} illustrates an application of ZO optimization to online sensor management.
 More applications are provided in Section\,\ref{sec: more_apps}.
 We  discuss open issues and state our conclusions in Sections\,\ref{sec: open_discussion} and
 \ref{sec: conclusion}, respectively.

\section{Gradient Estimation via ZO Oracle}
\label{sec: grad_est}

In this section, we provide an overview of gradient estimation techniques for optimization with a {black-box} objective function. The resulting gradient estimate forms the basis for constructing the descent direction used in ZO optimization algorithms. 
We categorize  the ZO gradient estimates into two types, 
\textit{$1$-point}, and \textit{multi-point} estimates, based on the   number of queried function evaluations. 
As the number of function evaluations increases, a more accurate gradient estimate is expected but at the cost of increased   query complexity.

\subsection{$1$-point estimate}
We start by the principles of \textit{randomized} gradient estimation in the context of      $1$-point estimation.
Let $f (\mathbf x)$ be a continuously differentiable objective function on a $d$-dimension variable $\mathbf x \in \mathbb R^d$. The $1$-point gradient estimate of $f$ has the  generic form  
\begin{align}\label{eq: grad_est_1}
   \hat{\nabla}f  (\mathbf x ) \Def \frac{\phi(d)}{\mu} f(\mathbf x + \mu   \mathbf u) \mathbf u,
\end{align}
where  $\mathbf u \sim p$ is a  random  direction vector drawn from a certain distribution $p$, which is typically chosen as either the standard multivariate normal distribution $\mathcal N(\mathbf 0, \mathbf I)$ \cite{nesterov2015random} or the multivariate uniform distribution $\mathcal U(\mathcal S(0,1))$ on a unit   sphere centered at $\mathbf 0$  with radius $1$ \cite{flaxman2005online}, $\mu > 0$ is a perturbation radius (also called a smoothing parameter), and $\phi(d)$ denotes  a certain dimension-dependent factor related to the  choice of the distribution  $p$.
If $p = \mathcal N(\mathbf 0, \mathbf I)$, then $\phi(d) = 1$; If $p = \mathcal U(\mathcal S(0,1))$, then $\phi(d) = d$.

The \textit{rationale} behind 
\eqref{eq: grad_est_1}
is that it is an \textit{unbiased} estimate of the gradient of 
the \textit{smoothed} version of $f$ 
over a random perturbation $\mathbf u \sim p^\prime$ with  smoothing parameter $\mu$, 
\begin{align}\label{eq: fmu}
   f_{\mu} (\mathbf x) \Def \mathbb E_{\mathbf u \sim p^\prime } [  f(\mathbf x + \mu \mathbf u)],
\end{align}
where $p^\prime$ is specified  as   $\mathcal N(\mathbf 0, \mathbf I)$ if $p = \mathcal N(\mathbf 0, \mathbf I)$ in \eqref{eq: grad_est_1}, or the multivariate uniform distribution on a unit ball $\mathcal U(\mathcal B(0,1))$ if $p = \mathcal U(\mathcal S(0,1))$ in \eqref{eq: grad_est_1}. 
The unbiasedness of  \eqref{eq: grad_est_1} with respect to $\nabla   f_{\mu} (\mathbf x)$ is  assured by
\cite{nesterov2015random,berahas2019theoretical}:
\begin{align}\label{eq: grad_est_smooth}
      \mathbb E_{\mathbf u \sim p} \left [   \hat{\nabla}f  (\mathbf x )  \right ] = \nabla f_{\mu} (\mathbf x) .
\end{align}
The meaning of \eqref{eq: grad_est_smooth} can be elucidated  by considering  the 
    scalar case $d = 1$.  
Given $p = \mathcal U(\mathcal S(0,1))$, applying the fundamental theorem of calculus to \eqref{eq: fmu} yields  $    \nabla f_{\mu} ( x)  = \frac{d}{dx} \int_{-\mu}^{\mu}  \frac{1}{2} f( x + u )   d u =  \frac{1}{2 \mu} [f( x + \mu) - f( x - \mu)]  $, which is equal to $\mathbb E_{ u \sim p}  [   \hat{\nabla}f  ( x )   ]$  from  \eqref{eq: grad_est_1}.

Although the $1$-point estimate \eqref{eq: grad_est_1} 
is unbiased with respect to the gradient of the smoothed  function $\nabla f_{\mu}(\mathbf x)$, it is a \textit{biased} approximation of the true gradient $\nabla f(\mathbf x)$. 
Furthermore,
the $1$-point estimate is not commonly used in practice since it suffers from  high variance, defined as $\mathbb E [\| \hat \nabla f(\mathbf x) - \nabla f_{\mu}(\mathbf x)  \|_2^2]$, which slows convergence  \cite{flaxman2005online}.


\subsection{Multi-point ZO estimate}

A natural extension of \eqref{eq: grad_est_1} is
the directional derivative approximation (\textit{$2$-point estimate}) \cite{duchi2015optimal,nesterov2015random},
\begin{align}\label{eq: grad_est_2}
   \hat{\nabla}f  (\mathbf x ) \Def \frac{\phi(d)}{\mu}  [ f(\mathbf x + \mu   \mathbf u) - f(\mathbf x) ] \mathbf u,
\end{align}
which  
satisfies the unbiasedness condition    \eqref{eq: grad_est_smooth} for any $\mathbf u$  such that  $\mathbb E_{\mathbf u \sim p} [ \mathbf u] = \mathbf 0$. 
The mean squared approximation error  of the gradient estimate \eqref{eq: grad_est_2} with respect to the true gradient $\nabla f (\mathbf x)$ obeys   \cite{liu2018_NIPS,berahas2019theoretical},
{\small \begin{align}
    \mathbb E [\| \hat{\nabla} f(\mathbf x) - \nabla f(\mathbf x) \|_2^2] =  O  ( d  ) \| \nabla f(\mathbf x) \|_2^2  + O \left ( \frac{\mu^2 d^3 + \mu^2 d}{\phi(d)}  \right ), 
    \label{eq: error_est_2}
\end{align}}%
where we adopt the big $O$ notation to highlight the dominant factors $d$ and $\mu$ affecting  gradient estimation error. 
It is worth noting that the (coordinate-wise) two-point  ZO estimate for finding the optimum of a regression function was initially proposed in the $1950$s \cite{kiefer1952stochastic}. 
This gradient estimation technique was further studied in the $1980$s in the context of simultaneous perturbation stochastic approximation \cite{spall1987stochastic,spall1992multivariate}.


The approximation error \eqref{eq: error_est_2} of the $2$-point estimate in \eqref{eq: grad_est_2} provides several insights. 
First, the gradient estimate gets better   as the smoothing parameter $\mu$ becomes smaller.  
 However, in a practical system, if $\mu$ is too small, then the function difference could be dominated by  system noise and it may fail to represent the  differential \cite{lian2016comprehensive,liu2018_NIPS}. 
 Thus, careful selection of the smoothing parameter $\mu$ is  important for  convergence   of ZO optimization methods. 
Second, 
different from the first-order stochastic gradient estimate,  the ZO gradient estimate yields a    dimension-dependent variance that increases as $ O  ( d  ) \| \nabla f(\mathbf x) \|_2^2 $. Thus, variance cannot be reduced even if $\mu \to 0$. 
Thus, some recent work has focused on the design of variance-reduced gradient estimates. 

Mini-batch sampling is the most commonly-used approach to reduce the variance of ZO gradient estimates \cite{duchi2015optimal,liu2017zeroth}.
Instead of using a single random direction, the average of $b$ i.i.d.  samples $\{ \mathbf u_i \}_{i=1}^b$ {drawn from $p$} are used for gradient estimation, leading to the \textit{multi-point estimate}
 \begin{align}\label{eq: grad_est_q}
   \hat{\nabla}f  (\mathbf x ) \Def \frac{\phi(d)}{\mu} \sum_{i=1}^b [ ( f(\mathbf x + \mu   \mathbf u_i) - f(\mathbf x) )  \mathbf u_i ],
\end{align}
with the approximation error  \cite{berahas2019theoretical}
{\small \begin{align}
 O  \left ( \frac{d}{b}  \right ) \| \nabla f(\mathbf x) \|_2^2  + O \left ( \frac{\mu^2 d^3 }{\phi(d) b}  \right ) + O \left ( \frac{\mu^2 d }{\phi(d) }  \right ). \label{eq: err_q_est}
\end{align}}%
In \eqref{eq: err_q_est}, the first two terms 
correspond to the reduced variance of  the $2$-point estimate $\mathbb E [\| \hat \nabla f(\mathbf x) - \nabla f_{\mu}(\mathbf x)  \|_2^2]$ due to the drawing of $b$ random direction vectors. And the third term,   
independent of $b$, corresponds to the approximation error due to the gradient of the smoothed function $\| \nabla f(\mathbf x) - \nabla f_{\mu}(\mathbf x) \|_2^2$.

When the number of function evaluations reaches the problem dimension $d$ in  \eqref{eq: grad_est_q}, then instead of using \textit{randomized}  directions $\{ \mathbf u_i \}_{i=1}^d$, one can  employ the \textit{deterministic coordinate-wise} gradient estimate 
$
\frac{1}{\mu}\sum_{i = 1}^d 
     [ ( f ( \mathbf x + \mu \mathbf e_i ) - f ( \mathbf x ) ) \mathbf e_i ] 
$, which yields a lower approximation error, of order $O(d \mu^2)$
\cite{kiefer1952stochastic,lian2016comprehensive,berahas2019theoretical}. Here $\mathbf e_i \in \mathbb R^d$ denotes the $i$th elementary basis vector, with $1$ at the $i$th coordinate and $0$s elsewhere. In practice, the multi-point gradient estimate \eqref{eq: grad_est_q} is usually implemented with $2 \leq b \leq d$. 
The previously introduced multi-point estimates are computed by using \textit{forward} differences of   function values. An alternative is the  \textit{central} difference   variant that uses   $ ( f(\mathbf x + \mu \mathbf u) - f(\mathbf x - \mu \mathbf u) ) $, where $\mathbf u$ can be either 
randomized or deterministic   \cite{kiefer1952stochastic,spall1987stochastic}.  
These central difference estimates have
similar   approximation errors to the forward difference estimates \cite{shamir2017optimal,lian2016comprehensive,berahas2019theoretical}.

\section{ZO Optimization Algorithms}
\label{sec: ZO_algs}

In this section, we present a unified algorithmic framework covering many commonly-used ZO optimization methods. We     provide  a  thorough overview of   existing algorithms   in   different   problem   settings   and   delve into the  factors  that  influence  their convergence. 

\subsection{The generic form of the ZO algorithm}
Let us
consider a stochastic optimization  problem  
 \begin{align}\label{eq:problem}
\min_{\mathbf x \in \mathcal X} ~ f(\mathbf x) \Def \mathbb{E}_{\boldsymbol \xi}[f(\mathbf x; \boldsymbol{\xi })],
\end{align}
where $\mathbf x \in \mathbb R^d$ are optimization variables, $\mathcal X$ is a closed convex  set, $f$ is a  possibly  \textit{nonconvex} objective function, 
and $\boldsymbol \xi$ is a certain random variable that captures
stochastic data samples or noise. 
If $\boldsymbol \xi$ 
obeys a uniform distribution over $n$ empirical   samples $\{ \boldsymbol \xi_i  \}_{i=1}^n$, then    problem \eqref{eq:problem} reduces to a finite-sum  formulation with  objective function 
$f(\mathbf x) = \frac{1}{n}\sum_{i=1}^n f(\mathbf x; \boldsymbol{\xi}_i)$. And if $\mathcal X = \mathbb R^d$, then problem \eqref{eq:problem} simplifies to the unconstrained optimization problem. 

\begin{algorithm}[htb]
\caption{Generic form of ZO optimization}
\begin{algorithmic}
  \STATE \textbf{Initialize} $\mathbf x_0 \in \mathcal X$, 
  gradient estimation operation $\phi(\cdot)$, descent  direction updating  operation $\psi(\cdot)$,
 number of iterations $T$, and learning rate $\eta_t > 0$ at iteration $t$,
\FOR{$t =  1,2,\ldots, T$}
\STATE  \textbf{1. Gradient estimation:} 
\begin{align}
    \hat{\mathbf g}_{t} = \phi ( \{ f(\mathbf x_t; \boldsymbol{\xi}_j) \}_{j \in  \Omega_t}^{t}), \label{eq: grad_est_alg}
\end{align}
where   $\Omega_t$ denotes a set of  mini-batch   stochastic samples used at iteration $t$,
\STATE \textbf{2. Descent direction computation:} 
\begin{align}
\mathbf m_{t} =  \psi(\{ \hat{\mathbf g}_i \}_{i=1}^t),    \label{eq: descent_alg}
\end{align}
\STATE \textbf{3. Point   updating:} 
\begin{align}
\mathbf x_{t} = \Pi_{\mathcal X} \left ( \mathbf x_{t-1}, \mathbf m_t, \eta_t
\right),    \label{eq: point_update_alg}
\end{align}
where $ \Pi_{\mathcal X}$ denotes a point updating operation subject to the constraint $\mathbf x \in \mathcal X$.
  \ENDFOR
\end{algorithmic}
\label{alg: general_form}
\end{algorithm}

Most ZO optimization methods mimic their first-order counterparts, and  involve three   steps, shown in Algorithm\,\ref{alg: general_form},  \textit{gradient estimation} \eqref{eq: grad_est_alg},  \textit{descent direction computation} \eqref{eq: descent_alg}, and  \textit{point updating} \eqref{eq: point_update_alg}. 
Without loss of generality, we specify \eqref{eq: grad_est_alg} as a variant of \eqref{eq: grad_est_q} built on a  mini-batch of empirical samples $\{ \boldsymbol{\xi}_{j}\}_{j \in \Omega_t}$,
\begin{align} \label{eq: grad_est_alg_specific}
\hat{ \mathbf g}_t  = \phi ( \{ f(\mathbf x_t; \boldsymbol{\xi}_j) \}_{j \in  \Omega_t}^{t},  \boldsymbol{\alpha}) = \frac{1}{| \Omega_t |} \sum_{j \in  \Omega_t} \hat{\nabla} f(\mathbf x_t; \boldsymbol{\xi}_j),
\end{align}
where $ \hat{\nabla} f(\mathbf x_t; \boldsymbol{\xi}_j)$ is given by \eqref{eq: grad_est_q} as   
the gradient of the function $f(\cdot; \boldsymbol{\xi})$, and $| \Omega_t |$ denotes the cardinality of the  set of mini-batch   samples at iteration $t$. 

Next, 
we elaborate on the   descent direction computation  and the point updating step   used in many ZO algorithms.

\subsubsection{ZO algorithms for unconstrained optimization}
We consider the ZO stochastic gradient descent (ZO-SGD) method \cite{ghadimi2013stochastic},  the  ZO   sign-based SGD (ZO-signSGD) \cite{liu2018signsgd}, the ZO stochastic variance reduced gradient   (ZO-SVRG) method \cite{gu2016zeroth,liu2018_NIPS,liu2018stochastic,JWL19}, and the
 ZO Hessian-based (ZO-Hess) algorithm  \cite{ye2018hessian,balasubramanian2019zeroth}. 
These algorithms   
employ the same point updating rule \eqref{eq: point_update_alg},
\begin{align*}
    \mathbf x_t = \mathbf x_{t-1} - \eta_t \mathbf m_t.
\end{align*}
However, they adopt different strategies to form the descent direction  $\mathbf m_t$ in  \eqref{eq: descent_alg}.

$\bullet$ {ZO-SGD \cite{ghadimi2013stochastic}:} The descent direction $\mathbf m_t$ is set as the current gradient estimate $
\mathbf m_t = \hat{\mathbf g}_t
$. Note that ZO-SGD  becomes  the ZO stochastic coordinate descent (ZO-SCD) method \cite{lian2016comprehensive} as the coordinate-wise gradient estimate is used.
Moreover, if  the full batch of stochastic samples are used, then ZO-SGD becomes ZO gradient descent (ZO-GD) \cite{nesterov2015random}.

$\bullet$ {ZO-signSGD \cite{liu2018signsgd}:} The descent direction $\mathbf m_t$ is given by the   sign  of  the current gradient estimate $
\mathbf m_t = \mathrm{sign}(\hat{\mathbf g}_t)
$, where $ \mathrm{sign} (\cdot)$ denotes the element-wise sign operation.
Using the sign operation scales down the   (coordinate-wise)
   estimation errors \cite{liu2018signsgd, cheng2019sign}. 

$\bullet$ ZO-SVRG  \cite{gu2016zeroth,liu2018_NIPS,liu2018stochastic,JWL19}: The descent direction $\mathbf m_t$ is formed by
 combining  $\hat{\mathbf g}_t$ with a control variate of reduced variance,
 $\mathbf m_t = \hat{\mathbf g}_t - \mathbf c_t + \mathbb E_{\boldsymbol{\xi}} [\mathbf c_t]$, where $\mathbf c_t$ denotes a control variate, which is commonly given by a  gradient estimate evaluated at $\mathbf x_{t-1}$ but the entire dataset of $n$ empirical samples. 
 
$\bullet$ ZO-Hess \cite{ye2018hessian}: The descent direction $\mathbf m_t$ incorporates the   approximate Hessian $\hat{\mathbf H}_t$ \cite{ye2018hessian},
 $\mathbf m_t = \hat{\mathbf H}_t^{-1/2}\hat{\mathbf g}_t   $, 
where $\hat{\mathbf H}_t$ is constructed  either by the second-order Gaussian Stein's identity \cite{balasubramanian2019zeroth} or the diagonalization-based Hessian approximation \cite{ye2018hessian}. The former approach was used in  \cite{balasubramanian2019zeroth} to develop the ZO stochastic cubic regularized Newton (ZO-SCRN) method.



\subsubsection{ZO algorithms for constrained optimization}
We next present the ZO projected SGD (ZO-PSGD) \cite{ghadimi2016mini}, the ZO stochastic mirror descent (ZO-SMD) \cite{duchi2015optimal}, the ZO stochastic conditional gradient (ZO-SCG) algorithm \cite{balasubramanian2019zeroth,balasubramanian2018zeroth}, and the ZO adaptive momentum method (ZO-AdaMM) \cite{chen2019zo}  for constrained optimization.
The aforementioned algorithms, with the exception of the ZO-AdaMM, specify the descent direction   \eqref{eq: descent_alg}  as the current gradient estimate $
\mathbf m_t = \hat{\mathbf g}_t
$. Their key difference  lies in how they implement the point updating step \eqref{eq: point_update_alg}.

$\bullet$ ZO-PSGD \cite{ghadimi2016mini}: By letting $\Pi_{\mathcal X}$ be the Euclidean distance based projection operation, the point updating step \eqref{eq: point_update_alg} is given by
$
\mathbf x_t = \argmin_{\mathbf x \in \mathcal X} \| \mathbf x - (\mathbf x_{t-1} - \eta_t \mathbf m_t )\|_2^2
$.

$\bullet$ ZO-SMD \cite{duchi2015optimal}: Upon defining a  Bregman divergence $D_{h}(\mathbf x, \mathbf y)$ with respect to a strongly convex and differentiable function $h$, 
$D_{h}(\mathbf x, \mathbf y) = h(\mathbf x) - h(\mathbf y) - (\mathbf x - \mathbf y)^T \nabla h(\mathbf y) $, the point updating step \eqref{eq: point_update_alg} is given by
$
\mathbf x_t = \argmin_{\mathbf x \in \mathcal X}  \mathbf m_t^T \mathbf x + \frac{1}{\eta_t} D_{h}(\mathbf x, \mathbf x_t) 
$. For example, if $h(\mathbf x) = \frac{1}{2}\| \mathbf x\|_2^2$, then $D_{h}(\mathbf x, \mathbf y) = \frac{1}{2} \| \mathbf x - \mathbf y \|_2^2$, and $\mathbf x_t = \argmin_{\mathbf x \in \mathcal X} \| \mathbf x - (\mathbf x_{t-1} - \eta_t \mathbf m_t )\|_2^2$, which reduces to ZO-PSGD.

$\bullet$ ZO-SCG \cite{balasubramanian2019zeroth,balasubramanian2018zeroth}: The point updating step \eqref{eq: point_update_alg} calls for a linear minimization oracle \cite{balasubramanian2019zeroth}, 
$\mathbf z_t = \argmin_{\mathbf x \in \mathcal X} \mathbf m_t^T \mathbf x$, and forms a feasible point update through the linear combination 
$\mathbf x_t = (1-\eta_t) \mathbf x_{t-1} + \eta_t \mathbf z_t$.
Similar algorithms,   known as ZO    Frank-Wolfe, were also developed in
\cite{chen2018frank,sahu2018towards}.

$\bullet$ ZO-AdaMM \cite{chen2019zo}: 
Different from ZO-PSGD, ZO-SMD and ZO-SCG, ZO-AdaMM adopts a momentum-type descent direction (rather than the current estimate  $\hat{\mathbf g}_t$), an adaptive learning rate (rather than the  constant rate $\eta_t$), and a projection operation under Mahalanobis distance (rather than Euclidean distance). 
ZO-AdaMM can strike  a  balance between the convergence speed and  accuracy. However, it requires tuning extra algorithmic hyperparameters in addition to the learning rate and smoothing parameters \cite{chen2019zo,chen2018convergence}. 


\subsection{ZO optimization in  complex settings}
Here we   review ZO algorithms for
{composite optimization}, {min-max optimization}, {distributed optimization}, and  structured high-dimensional optimization.

\subsubsection{ZO composite optimization} 
Consider the following   problem, with a  smooth+nonsmooth composite objective function,
\begin{align}\label{eq: prob_composite}
    \begin{array}{ll}
\displaystyle \minimize_{\mathbf x \in \mathbb R^d}         &  f(\mathbf x) + g(\mathbf x),
    \end{array}
\end{align}
where $f$ is a {black-box} smooth function (possibly nonconvex), and $g$ is a white-box non-smooth regularization function. The form of problem   \eqref{eq: prob_composite}   arises in many sparsity-promoted applications, e.g., adversarial attack generation   \cite{Zhao_2019_ICCV} and online sensor management \cite{liu2017zeroth}.
 The ZO   proximal SGD (ZO-ProxSGD) algorithm \cite{ghadimi2016mini}
and   ZO (stochastic) alternating direction method of multipliers (ZO-ADMM) \cite{gao2014information,liu2017zeroth,huang2019zeroth} were developed to solve  problem \eqref{eq: prob_composite}.
We remark that problem \eqref{eq:problem} can also be cast as \eqref{eq: prob_composite} by introducing the indicator function of  the constraint $\mathbf x \in \mathcal X$ in the objective of \eqref{eq:problem} by letting $g(\mathbf x ) = 0$ if $\mathbf x \in \mathcal X$ and $\infty$ if $\mathbf x \notin \mathcal X$.

\subsubsection{ZO min-max optimization} 
 By \emph{min-max}, we mean that the problem is a composition of  inner maximization and outer minimization of the objective function,
\begin{align}\label{eq: prob_min_max}
    \begin{array}{ll}
\displaystyle \min_{\mathbf x \in \mathcal X} \max_{\mathbf y \in \mathcal Y}         &  f(\mathbf x, \mathbf y)
    \end{array}
\end{align}
where $\mathbf x \in \mathbb{R}^{d_x}$ and $\mathbf y \in \mathbb{R}^{d_y}$ are optimization variables (for ease of notation,  let $d_x = d_y = d$), $f$ is a black-box objective function, 
and  $\mathcal X $ and $\mathcal Y  $ are 
compact 
convex sets.  
One motivating application behind problem \eqref{eq: prob_min_max} is the design of \textit{black-box poisoning attack} \cite{liu2019min}, where   the attacker deliberately influences the training data (by injecting  poisoned samples) to manipulate the results of a black-box predictive model. To solve the problem  posed in \eqref{eq: prob_min_max},
the work \cite{liu2019min,wang2020zeroth} presented 
  efficient ZO min-max  algorithms for stochastic and deterministic  bi-level  optimization  with nonconvex outer minimization over $\mathbf x$ and strongly concave inner maximization over $\mathbf y$. They proved the  convergence rate to be sub-linear  when 
    gradient estimation is integrated with 
  alternating (projected) stochastic gradient descent-ascent methods.
    

\subsubsection{ZO distributed optimization}
Consider the minimization of  a network cost,  given by the sum of local objective functions $\{ f_i \}$ at multiple agents 
\begin{align}\label{eq: prob_distr}
    \begin{array}{ll}
\displaystyle \minimize_{\{ \mathbf x_i \in \mathcal X \} }         &  \sum_{i=1}^N f_i (\mathbf x_i) \\
       \st  & \mathbf x_i = \mathbf x_j, ~ \forall j \in \mathcal N (i).
    \end{array}
\end{align}
Here $\mathcal N(i)$ denotes the set of neighbors of agent/node $i$, and the underlying network/graph is  connected,
namely, there exists a path between
every pair of distinct nodes. 
Some recent works  have   started to tackle the distributed   optimization problem \eqref{eq: prob_distr} with black-box objectives.
In \cite{sahu2018distributed}, a distributed  Kiefer Wolfowitz type  ZO algorithm was proposed  along with  convergence analysis, for the case that the objective functions $\{f_i\}$  are  strongly convex.
In \cite{yuan2015zeroth,yu2019distributed}, the ZO distributed (sub)gradient algorithm and the ZO distributed mirror descent algorithm were developed for nonsmooth convex optimization. 
In \cite{hajinezhad2017zeroth,tang2019distributed}, the convergence of consensus-based distributed ZO algorithms was established for
   noncovnex (unconstrained)  optimization.










\subsubsection{Structured high-dimensional optimization}
Compared to FO algorithms, ZO algorithms typically suffer from  a slowdown (proportional to the problem size $d$) in convergence. Thus, some recent works attempt to mitigate this limitation when  solving high-dimensional (large $d$) problems. 
The work \cite{wangdu18} explored the functional sparsity structure, under which the objective function $f$ depends only on a subset of $d$ coordinates.   This assumption also implies the 
  gradient sparsity, which enabled the development of  a   LASSO based algorithm for  gradient estimation, and eventually yielded poly-logarithmic dependence on $d$ when $f$ is convex. 
And
  the work \cite{balasubramanian2019zeroth} 
established the convergence rate of ZO-SGD  which   depends on $d$ only
poly-logarithmically under the assumption of gradient sparsity. 
In addition, the work  \cite{golovin2019gradientless}  proposed  a direct search based algorithm, which  yields the convergence rate that is  poly-logarithmically dependent on dimensionality for any monotone transform of a smooth and strongly convex
objective with a low-dimensional structure. That is, the objective function $f(\mathbf x)$ is supported on a low dimensional manifold $\mathcal X$.
Another work \cite{li2020zeroth} studied the problem of ZO optimization  on Riemannian manifolds, and proposed  algorithms that
only depend on the intrinsic
dimension of the manifold by using 
  ZO Riemannian gradient estimates.





\begin{table*}[htb]
{\centering
\caption{Comparison of different ZO algorithms in problem setting,  gradient estimation,  smoothing parameter,    convergence error, and function query complexity.}
\label{table: SZO_complexity_T}
\begin{adjustbox}{max width=1\textwidth }
\begin{threeparttable}
\begin{tabular}{|c|c|c|c|c|c|}
\hline
\textbf{Method}        & \begin{tabular}[c]{@{}c@{}} \textbf{Problem} \\ \textbf{structure}\end{tabular}       & \begin{tabular}[c]{@{}c@{}} \textbf{Gradient}\\\textbf{estimation}\end{tabular} 
&     \begin{tabular}[c]{@{}c@{}} \textbf{Smoothing}\\ \textbf{parameter $\mu$}   \end{tabular} 
&  \begin{tabular}[c]{@{}c@{}} \textbf{Convergence error} \\  ($T$ iterations)  \end{tabular}       
& \begin{tabular}[c]{@{}c@{}} \textbf{Query complexity} \\ ($T$ iterations) \end{tabular} 
\\ \hline  
ZO-GD  \cite{nesterov2015random}
& NC, UnCons$^1$ 
&  $2$-point {GauGE$^2$}
&  $ O\left (\frac{1}{\sqrt{dT}} \right )$ 
& 
$O\left (\frac{d}{T} \right )$ 
&
$O\left (|\mathcal D|  T \right )^3 $ 
\\ \hline
ZO-SGD  \cite{ghadimi2013stochastic} & NC, UnCons &  $2$-point GauGE
& $O \left (\frac{1}{d\sqrt{T}}\right )$   
&
$O\left ( \frac{\sqrt{d}}{\sqrt{T}}  \right ) $ 
& $O\left (  T \right ) $ 
\\ \hline
ZO-SCD \cite{lian2016comprehensive} & NC, UnCons &  $2$-point CooGE$^2$
& $O \left (\frac{1}{\sqrt{T}} + \frac{1}{(dT)^{1/4}} \right )$  
& 
$O \left ( \frac{\sqrt{d}}{\sqrt{T}} \right )$  & $O\left (  T \right ) $ 
\\ \hline
ZO-signSGD \cite{liu2018signsgd}  & NC, UnCons  &  $b$-point UniGE$^2$
& $O \left ( \frac{1}{\sqrt{dT}} \right )$
&  $O\left (\frac{\sqrt{d}}{\sqrt{T}} + \frac{\sqrt{d}}{\sqrt{b}}   \right )^4$ &
$O\left ( b T \right ) $ 
\\ \hline 
ZO-SVRG \cite{liu2018_NIPS} &
NC, UnCons
&
$b$-point UniGE   
&
$O\left ( \frac{1}{\sqrt{dT}}  \right )$
&
$ O\left (\frac{{d}}{{T}} + \frac{1}{b}  \right )$ 
& 
\begin{tabular}[c]{@{}c@{}} $O \left ( |\mathcal D| s + b s m \right )$ \\ $T = sm$ \end{tabular}
\\ \hline
ZO-Hess \cite{ye2018hessian} &
SC, UnCons
&
$b$-point GauGE  
&
$O\left ( \frac{1}{d}  \right )$
&
$ O\left ( e^{-bT/d} \right )$ 
& 
$O \left ( bT \right )$
\\ \hline
\begin{tabular}[c]{@{}c@{}} ZO-ProxSGD /  \\
ZO-PSGD
\cite{ghadimi2016mini} 
\end{tabular}  &
NC, Cons
&
$b$-point GauGE 
&
$ O\left ( \frac{1}{\sqrt{dT}}\right )$
&
$ O \left ( 
\frac{d^2}{b  T} + \frac{d}{b } 
\right )
$ &  
$O\left (b T \right ) $ 
\\ \hline
ZO-SMD \cite{duchi2015optimal}&
C, Cons
&
$2$-point GauGE
& $O\left ( \frac{ 1}{ d t }\right )$
 & $ O\left ( \frac{ \sqrt{d}}{\sqrt{T} }\right )$ & $O(T)$
\\ \hline
ZO-SCG \cite{balasubramanian2019zeroth} &
NC, Cons 
&
$b$-point GauGE  
&
$O\left ( \frac{ 1}{\sqrt{d^3 T}}\right )$
& $  O \left ( \frac{1}{\sqrt{T}} + \frac{d \sqrt{T}}{b}   \right )$
& $O\left ( b T \right ) $ 
\\ \hline
ZO-AdaMM \cite{chen2019zo} &
NC, Cons 
&
$b$-point GauGE  
&
$O\left ( \frac{ 1}{\sqrt{d T}}\right )$
& $  O \left ( \frac{d}{T} + \frac{d  }{b}   \right )$
& $O\left ( b^2 T \right ) $ 
\\ \hline
ZO-ADMM \cite{liu2017zeroth} &
C, Composite
&
$b$-point UniGE   
&
$O\left ( \frac{1}{d^{1.5}t}\right )$ 
&
$   O \left ( \frac{\sqrt{d}}{\sqrt{ bT }} \right )$ &  $O\left ( b T \right ) $ 
\\ \hline
ZO-Min-Max \cite{liu2019min} &
NC, Cons
&
$b$-point UniGE   
&
$O\left ( \frac{1}{d \sqrt{T} }\right )$ 
&
$   O \left (  \frac{1}{T} + \frac{d}{b}  \right )$ &  $O\left ( b^2 T \right ) $ 
\\ \hline
Dist-ZO \cite{tang2019distributed} &
NC, UnCons
&
$2$-point UniGE   
& $\frac{1}{\sqrt{dt}}$
&
$   O \left (  \frac{\sqrt{d}}{\sqrt{T}}     \right )$ &  $O\left (  T \right ) $ 
\\ \hline
ZO-SCRN \cite{balasubramanian2019zeroth} &
NC, UnCons
&
$b$-point UniGE   
& $\frac{1}{d^{5/2}T^{1/3}}$
&
\begin{tabular}[c]{@{}c@{}} $   O \left (  \frac{1}{T^{4/3}}     \right )$  \\
$b = O \left ( d T^{4/3}\right )$
\end{tabular}
&  
\begin{tabular}[c]{@{}c@{}} $O\left ( b T \right ) $  
\end{tabular}
\\ \hline
\end{tabular}
\begin{tablenotes}
    \small
    \item[1] Problem setting: NC, C,   SC, UnCons, Cons and Composite represent  nonconvex, convex, strongly convex, unconstrained, constrained and composite optimization respectively.\\
    \item[2] GauGE and UniGE represent the   gradient estimates   using random direction vectors drawn from $\mathcal N(\mathbf 0, \mathbf I)$ and $\mathcal U(\mathcal S(0,1))$, respectively. 
    CooGE represents the coordinate-wise partial derivative estimate.\\
    \item[3] $\mathcal D$ denotes the entire dataset.
\end{tablenotes}
\end{threeparttable}
\end{adjustbox}
  }
\end{table*}


\subsection{Convergence rates}

We first  elaborate on   the criteria used to analyze the convergence rate of ZO algorithms under different problem settings. 

 \textit{1) Convex optimization:} The convergence error is measured by the \textit{optimality gap of function values} $ \mathbb E \left [ f(\mathbf x_T) - f(\mathbf x^*) \right ]$ for a convex objective $f$, where $\mathbf x_T$ denotes the   updated point at  the final iteration $T$,   $\mathbf x^*$ denotes the optimal solution, and  the expectation is taken over the full probability space, e.g., random gradient approximation and stochastic sampling.

 \textit{2) Online convex optimization:} The \textit{cumulative regret} \cite{hazan2016introduction} is typically used in place of the optimality gap, namely, 
$\mathbb E \left [ \sum_{t = 1}^T f_t(\mathbf x_t) - \min_{\mathbf x } \sum_{t = 1}^T f_t(\mathbf x) \right ] $ for an online convex cost function $f_t$, e.g., $f_t(\cdot) =  f(\cdot; \boldsymbol{\xi}_t)$ in problem \eqref{eq:problem}. 

\textit{3) Unconstrained nonconvex optimization:} The convergence is evaluated by the first-order stationary condition in terms of  
the \textit{squared gradient norm}  $\frac{1}{T} \sum_{t=1}^T \mathbb E [ \| \nabla f(\mathbf x_t) \|_2^2 ]$ for the nonconvex objective $f$.
Since first-order stationary points could be saddle points of a nonconvex optimization problem,
the \textit{second-order stationary condition} is also  used to ensure the local optimality of a first-order stationary point (namely, escaping saddle-points) \cite{balasubramanian2019zeroth,vlatakis2019efficiently}.
The work \cite{balasubramanian2019zeroth} and \cite{vlatakis2019efficiently}
focused on stochastic optimization and deterministic optimization, respectively.


  \textit{4) Constrained nonconvex optimization:}
The criterion for convergence is commonly determined by detecting a  sufficiently small squared norm of the
\textit{gradient mapping}   \cite{reddi2016proximal,ghadimi2016mini},   
 $
  P_{\mathcal X} (\mathbf x_t,  \nabla f(\mathbf x_t), \eta_t ) \Def  \frac{1}{\eta_t} \left[ \mathbf x_t -\Pi_{\mathcal X} \left ( \mathbf x_t - \eta_t \nabla f(\mathbf x_t) \right ) \right ]
 $, where    the notation follows \eqref{eq: point_update_alg}. 
$ P_{\mathcal X} (\mathbf x_t,  \nabla f(\mathbf x_t), \eta_t )$ can naturally  be interpreted as
the  projected gradient, which offers a feasible update  from the previous point $\mathbf x_t$. 
The \textit{Frank-Wolfe duality gap} is another commonly-used   convergence criterion    \cite{balasubramanian2019zeroth,balasubramanian2018zeroth,chen2018frank,sahu2018towards}, given by $\max_{\mathbf x \in \mathcal X} \inp{\mathbf x - \mathbf x_t}{ -\nabla f(\mathbf x_t)}$. It is always non-negative, and becomes $0$ if and only if $\mathbf x_t \in \mathcal X$ is a stationary point.
 
 More generally,
 given a convergence measure $\mathcal M(\cdot)$,  $\mathbf x$ is called an $\epsilon$-optimal solution if $\mathcal M(\mathbf x) \leq \epsilon$.
 The convergence error is typically expressed as a function of the number of iterations $T$, relating the convergence rate to the iteration complexity. 
 Since the    convergence analysis  of existing ZO algorithms varies under different problem   domains and algorithmic parameter settings,   in Table\,\ref{table: SZO_complexity_T} we compare the convergence performance of    ZO algorithms covered in this section from $5$ perspectives: problem structure, type of gradient estimates,  smoothing parameter,    convergence error, and function query complexity.

\section{Application: Adversarial Example Generation}
\label{sec: adv}
In this section, we present 
the application of  ZO optimization to the generation of    prediction-evasive
\textit{adversarial examples} 
to fool DL models. 
Adversarial examples, also known as evasion attacks,  are inputs corrupted with imperceptible adversarial perturbations (to be designed) toward misclassification (namely, prediction different from true image labels)
\cite{goodfellow2015explaining,carlini2017towards}.

Most studies on adversarial vulnerability of DL have been  restricted to the \textit{white-box} setting where  the adversary has complete access and knowledge of the target system (e.g., DNNs) \cite{goodfellow2015explaining,carlini2017towards}. 
However, it is often the case that the internal states/configurations and the operating mechanism of  DL systems are not revealed to the practitioners (e.g., Google Cloud Vision API). This gives rise to the problem of \textit{black-box adversarial attacks} \cite{chen2017zoo,tu2018autozoom,ilyas2018blackbox,brendel2017decision,cheng2018query,zhao2020towards}, where 
 the only mode of interaction of the adversary with the system is via submission of inputs and receiving the corresponding predicted outputs.

\begin{table}[htb]
\caption{Comparison of various ZO methods for generating untargeted adversarial attacks against  Inception V3 model over $5$ ImageNet images. Row $1$: true labels of given images. Row $2$-$4$: Results obtained using ZO-SMD, which include perturbed images, corresponding prediction labels, $\ell_2$ norm of  perturbations, and the number of queries when achieving the first successful black-box attack. A similar explanation holds for other rows, except that different ZO methods are used. 
} \label{table:mnist_iteration_2}
\centering 
  \begin{adjustbox}{max width=0.98\textwidth}
  \begin{tabular}
      {     c       c      c     c    c    c   c}
     \toprule 
      	\begin{tabular}[c]{@{}c@{}} {True} {label:}\end{tabular} & {brambling} & {cannon}  & {pug-dog}  & {fly}  & {armadillo} 
      	& {balloon}  
      	\\
  \toprule       &  &   &    &   &  
      	&    \vspace*{-0.1in}  \\
\begin{tabular}[c]{@{}c@{}} {Perturbed image} \\ \textbf{(ZO-PSGD):}   \end{tabular}
	&
    \parbox[c]{4.25em}{\includegraphics[width=0.6in]{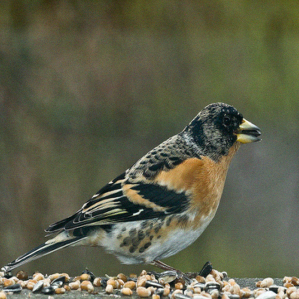}} &  
        \parbox[c]{4.25em}{\includegraphics[width=0.6in]{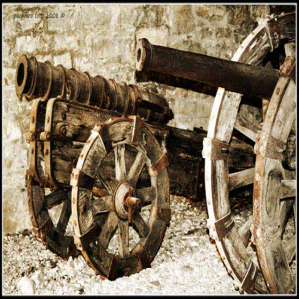}} &   
        \parbox[c]{4.25em}{\includegraphics[width=0.6in]{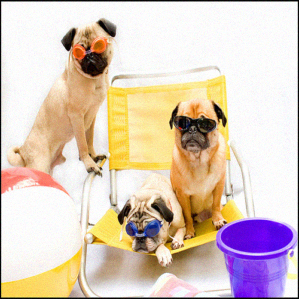}} &  
        \parbox[c]{4.25em}{\includegraphics[width=0.6in]{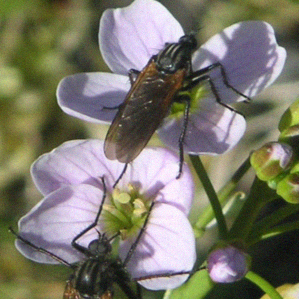}} &  
        \parbox[c]{4.25em}{\includegraphics[width=0.6in]{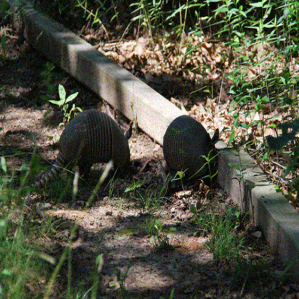}} & 
        \parbox[c]{4.25em}{\includegraphics[width=0.6in]{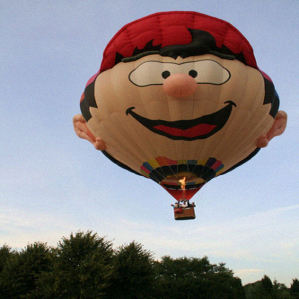}}  
        \\
        \midrule
        \begin{tabular}[c]{@{}c@{}}  {Prediction label:}
        \end{tabular}
        & 
        goldfinch 
        & plow 
        & bucket 
        & longicorn 
        & croquet ball & parachute  
        \\      
      \midrule
    	{$\ell_2$ Distortion:} & 35.2137 & 87.7304 & 72.3397 & 111.068 & 172.719 & 35.9330  
    	\\
    	\midrule
 \begin{tabular}[c]{@{}c@{}}   {\# of queries}:
        \end{tabular} & 7640 & 150 & 130 & 50 & 210 & 5830
        \\
      \midrule &&&&&&  \vspace*{-0.1in}  \\
      \begin{tabular}[c]{@{}c@{}} {Perturbed image} \\ \textbf{(ZO-SMD):}   \end{tabular}
      &
        \parbox[c]{4.25em}{\includegraphics[width=0.6in]{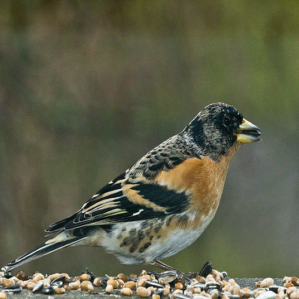}} &
      \parbox[c]{4.25em}{\includegraphics[width=0.6in]{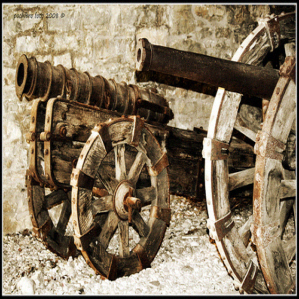}} & 
        \parbox[c]{4.25em}{\includegraphics[width=0.6in]{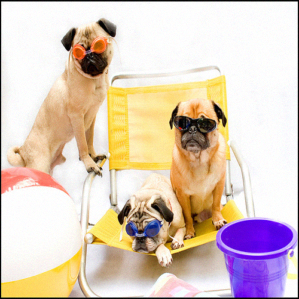}} &  
        \parbox[c]{4.25em}{\includegraphics[width=0.6in]{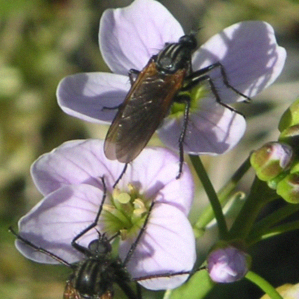}} &   
        \parbox[c]{4.25em}{\includegraphics[width=0.6in]{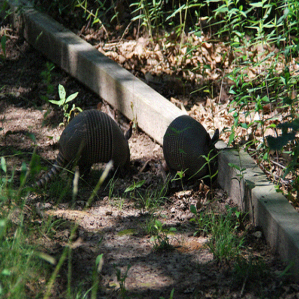}} &  
        \parbox[c]{4.25em}{\includegraphics[width=0.6in]{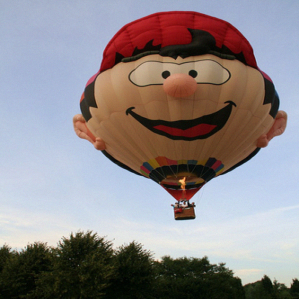}} 
        \\
        \midrule
        \begin{tabular}[c]{@{}c@{}}  {Prediction label:}
        \end{tabular} & goldfinch & plow & bucket & cicada & croquet ball & parachute
        \\      
       \midrule
        \begin{tabular}[c]{@{}c@{}}  {$\ell_2$ distortion:}
        \end{tabular} & 8.9708 & 26.0126 & 20.9504 & 30.0968 & 45.097 & 10.9023  
        \\
        \midrule
  \begin{tabular}[c]{@{}c@{}}   {\# of queries}:
        \end{tabular}  & 29980 & 350 & 260 & 140 & 570 & 15830
        \\
      \midrule &&&&&&  \vspace*{-0.1in}  \\
      \begin{tabular}[c]{@{}c@{}} {Perturbed image} \\ \textbf{(ZO-AdaMM):}   \end{tabular}
      &
          \parbox[c]{4.25em}{\includegraphics[width=0.6in]{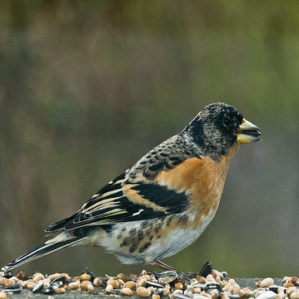}} &
        \parbox[c]{4.25em}{\includegraphics[width=0.6in]{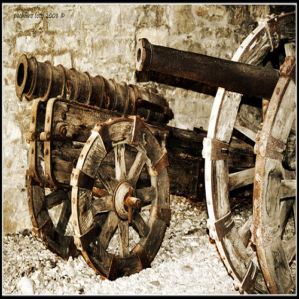}} &
        \parbox[c]{4.25em}{\includegraphics[width=0.6in]{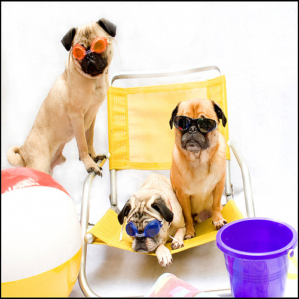}} &
        \parbox[c]{4.25em}{\includegraphics[width=0.6in]{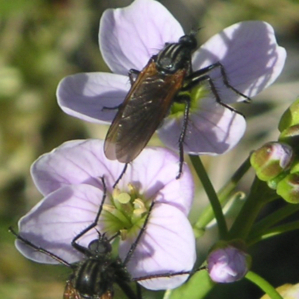}} &
        \parbox[c]{4.25em}{\includegraphics[width=0.6in]{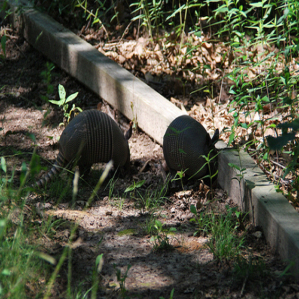}} &
        \parbox[c]{4.25em}{\includegraphics[width=0.6in]{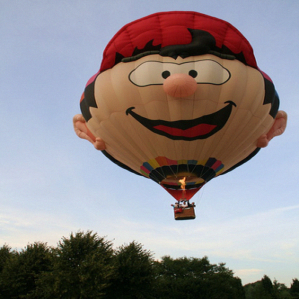}} 
        \\
        \midrule
        \begin{tabular}[c]{@{}c@{}}  {Prediction label:}
        \end{tabular} & goldfinch & plow & bucket & cicada & croquet ball & parachute
        \\    
\midrule
 \begin{tabular}[c]{@{}c@{}}  {$\ell_2$ distortion:}
        \end{tabular} & 8.0502 & 5.7359 & 4.5753 & 4.4456 & 6.3149 & 7.7405 
        \\
        \midrule
 \begin{tabular}[c]{@{}c@{}}   {\# of queries}:
        \end{tabular}  & 53300 & 1710 & 790 & 540 & 2780 & 32900
        \\
      \midrule &&&&&&  \vspace*{-0.1in}  \\
      \begin{tabular}[c]{@{}c@{}} {Perturbed image} \\ \textbf{(ZO-NES):}   \end{tabular}
      &
          \parbox[c]{4.25em}{\includegraphics[width=0.6in]{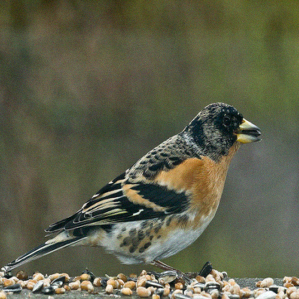}} &
        \parbox[c]{4.25em}{\includegraphics[width=0.6in]{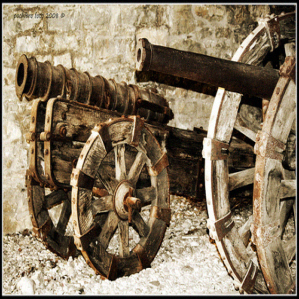}} &
        \parbox[c]{4.25em}{\includegraphics[width=0.6in]{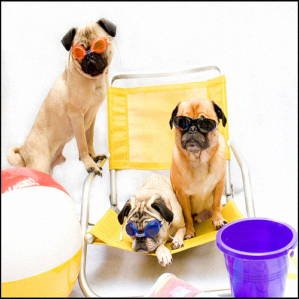}} &
        \parbox[c]{4.25em}{\includegraphics[width=0.6in]{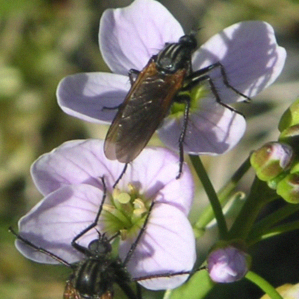}} &
        \parbox[c]{4.25em}{\includegraphics[width=0.6in]{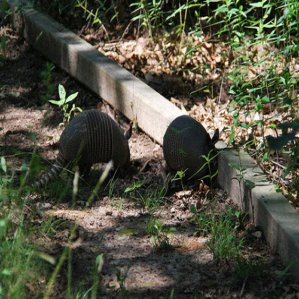}} &
        \parbox[c]{4.25em}{\includegraphics[width=0.6in]{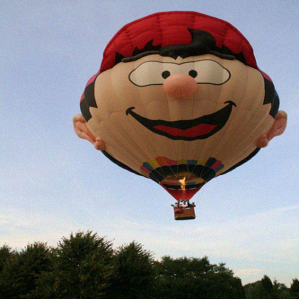}} 
        \\
       \midrule
        \begin{tabular}[c]{@{}c@{}}  {Prediction label:}
        \end{tabular} & goldfinch & plow & bucket & cicada & croquet ball & parachute 
        \\   
        \midrule
 \begin{tabular}[c]{@{}c@{}}  {$\ell_2$ distortion:}
        \end{tabular} & 54.9956 & 34.5852 & 28.7035 & 28.9158 & 40.1483 & 51.7116 
        \\
        \midrule
 \begin{tabular}[c]{@{}c@{}}   {\# of queries}:
        \end{tabular}  & 22110 & 1080 & 830 & 430 & 2280 & 15340
        \\
\bottomrule
  \end{tabular}
  \end{adjustbox}
\end{table} 


More formally, let  $\mathbf{z}$ denote a legitimate example, and $\mathbf z^\prime \Def \mathbf z+ \mathbf x$ denote an adversarial example with the adversarial perturbation $\mathbf x$. Given the learned ML/DL model $\boldsymbol{\theta}$, the problem of adversarial example generation can be cast as an optimization problem of the following generic form \cite{carlini2017towards,Zhao_2019_ICCV}
\begin{align}\label{eq: attack_general}
\begin{array}{ll}
    \displaystyle \minimize_{ \mathbf x \in \mathbb R^d } & f(\mathbf z + \mathbf x ; \boldsymbol{\theta}) + {\lambda} g( \mathbf x)
    \\
  \st    &   \| \mathbf x\|_\infty \leq \epsilon, ~\mathbf z^\prime \in [0,1]^d, 
\end{array}
\end{align}
where $f(\mathbf z^\prime ; \boldsymbol{\theta})$ denotes the  (black-box) attack loss function for fooling the model $\boldsymbol{\theta}$ using the perturbed input $\mathbf z^\prime$ (see \cite{chen2017zoo} for a specific formulation),  $g( \mathbf x ) $ is a regularization function that penalizes the sparsity or the structure of adversarial perturbations, e.g., group sparsity in \cite{xu2018structured}, $\lambda \geq 0$ is a regularization parameter,     the    $\ell_{\infty}$ norm  enforces  similarity between $\mathbf z^\prime$ and $\mathbf z$, and the input space of ML/DL systems  is normalized to $[0,1]^d$.  If $\lambda \neq 0$, then problem \eqref{eq: attack_general} is in the form of composite optimization, and   ZO-ADMM is a well-suited optimizer.  If $\lambda = 0$, a solution to problem \eqref{eq: attack_general} is known as a black-box $\ell_\infty$ attack \cite{ilyas2018blackbox}, which can be obtained using ZO methods for constrained optimization.

 In Table\,\ref{table:mnist_iteration_2}, we present   black-box $\ell_\infty$ attacks  with respect to $5$ ImageNet images  against the  Inception V3 model \cite{szegedy2016rethinking}.
 The adversarially perturbed images are obtained from $4$ ZO methods including
 ZO-PSGD, ZO-SMD, ZO-AdaMM, and ZO-NES (a projected version of ZO-signSGD that is used in practice \cite{ilyas2018blackbox}).
 We demonstrate the attack performance of different ZO algorithms in terms of the $\ell_2$ norm of the generated perturbations and the number of  queries needed to achieve a first successful black-box attack.
 As we can see, ZO-PSGD typically has the fastest speed of converging to a valid adversarial example, while ZO-AdaMM has the best convergence accuracy in terms of the smallest distortion required to fool the neural network.

\section{Application:  Online Sensor Management}
\label{sec: ZO_sensor}
 ZO optimization is also advantageous when 
it is \textit{difficult} to compute the first-order gradient of an objective function. Online sensor management provides an example of such a scenario \cite{chen2017bandit,liu2017zeroth}. 
Sensor selection for parameter estimation is a fundamental problem   in  smart grids, communication systems, and wireless sensor networks \cite{hero2011sensor}. The goal is to seek the optimal  tradeoff between   sensor activations and the estimation accuracy over a time period.

We consider the cumulative loss for online sensor selection  \cite{liu2017zeroth}
\begin{align}
\begin{array}{ll}
\displaystyle \minimize_{\mathbf x \in \mathbb R^d} & \displaystyle  \frac{1}{T}\sum_{t=1}^T \left [ - \mathrm{logdet} \left (\sum_{i=1}^d  x_i \mathbf a_{i,t} \mathbf a_{i,t}^T \right ) \right ] \\
\st & \mathbf 1^T \mathbf x = m_0, ~ \mathbf 0 \leq \mathbf x \leq 1,
\end{array}
\label{eq: prob_sensorSel_online}
\end{align}
 where $\mathbf x \in \mathbb R^d$ is the optimization variable,
 $d$ is the number of sensors, 
 $\mathbf a_{i,t} \in \mathbb R^n$ is the observation coefficient of sensor $i$ at time $t$, 
 and $m_0$ is the  number of selected sensors.  
The objective function of \eqref{eq: prob_sensorSel_online} can be interpreted as 
 the  log determinant of the error covariance matrix associated with the maximum likelihood estimator for parameter estimation \cite{joshi2009sensor}. 
The constraint $\mathbf 0 \leq \mathbf x \leq  \mathbf 1$ is a relaxed convex hull of the
 Boolean constraint   $\mathbf x \in \{0,1 \}^m$, which   encodes whether or not a sensor is selected.

Conventional   methods such as
  projected gradient (first-order)  and     interior-point (second-order) algorithms can be used to solve problem  \eqref{eq: prob_sensorSel_online}. However, both methods involve the calculation of inverses of large matrices that are
  necessary to evaluate
the gradient of  the   cost function. The matrix inversion step is usually a bottleneck while acquiring the gradient information in high dimensions and it is particularly problematic in the online optimization setting. 
Since problem \eqref{eq: prob_sensorSel_online} involves mixed equality and inequality constraints, it has been shown  \cite{liu2017zeroth} that ZO-ADMM  is an effective ZO optimization method for circumventing the computational bottleneck. 

\begin{figure}[htb]
	\centering
	\begin{tabular}{c}
	     \includegraphics[width=1\columnwidth]{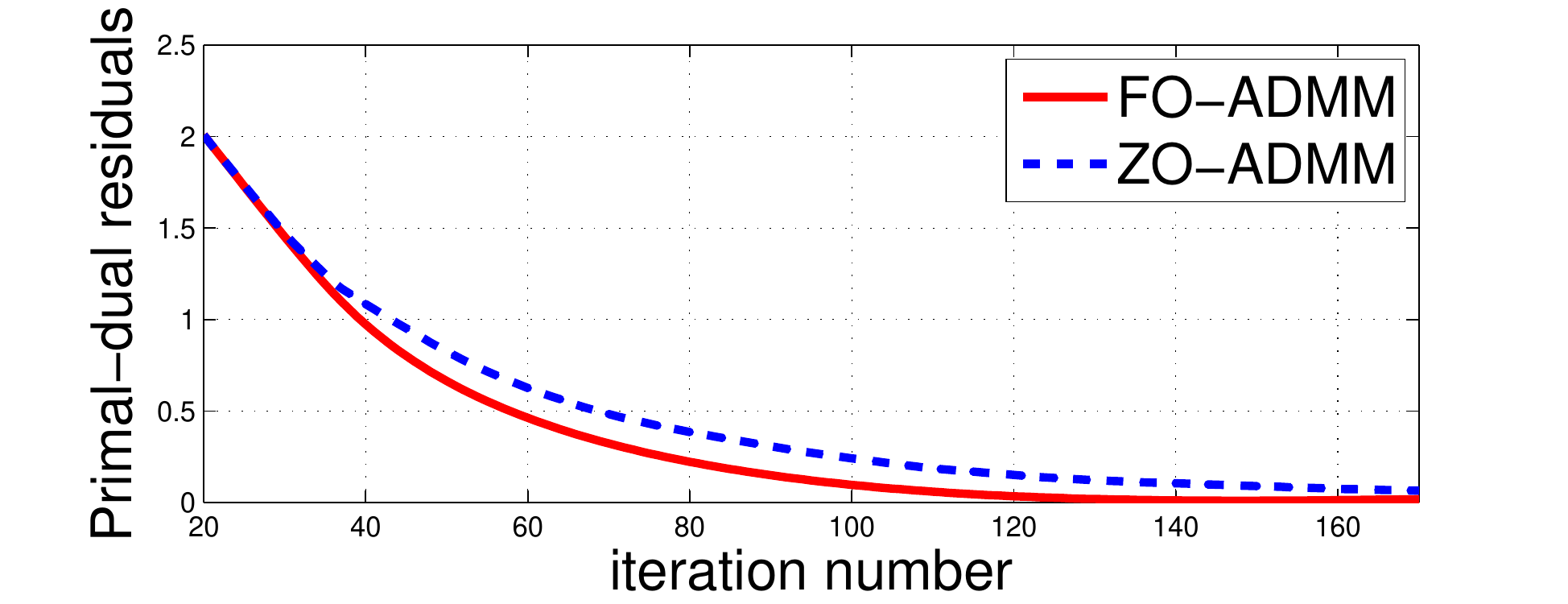}
  \\
	     \includegraphics[width=1\columnwidth]{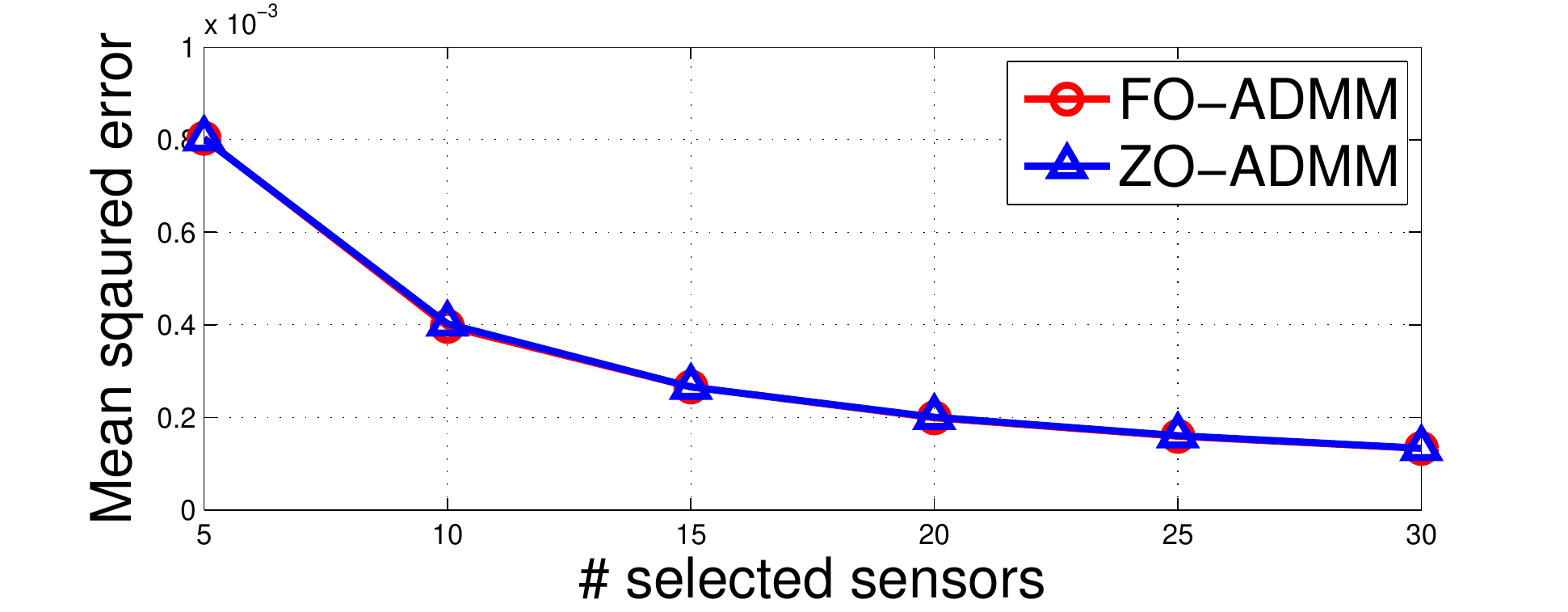}
	\end{tabular}
		\caption{Comparison between ZO-ADMM and FO-ADMM for solving the   sensor selection problem \eqref{eq: prob_sensorSel_online}. Top: ADMM primal-dual residuals versus number of iterations. Bottom: Mean squared error of activated sensors for parameter estimation versus the total number of selected sensors.
        }
		\label{fig: sensr}
\end{figure}

In Figure\,\ref{fig: sensr}, we compare the performance   of ZO-ADMM and that of FO-ADMM \cite{suzuki2013dual} for sensor selection. In the top plots, 
we demonstrate the primal-dual residuals in ADMM against the number of iterations. As we can see,  ZO-ADMM has a slower convergence rate than FO-ADMM, and it approaches the accuracy of FO-ADMM as the number of iterations   increases. In the bottom plots, we show
the mean squared error (MSE) of parameter estimation using 
 different number of selected sensors $m_0$ in \eqref{eq: prob_sensorSel_online}. 
As we can see, ZO-ADMM yields almost the  same MSE  as FO-ADMM in the context of parameter estimation using $m_0$ activated sensors, determined by the hard thresholding of   continuous sensor selection schemes, i.e., solutions of problem \eqref{eq: prob_sensorSel_online} obtained from ZO-ADMM  and FO-ADMM. 

\section{Other  Recent Applications}
\label{sec: more_apps}
In this section we discuss some  other recent applications of ZO optimization in signal processing and machine learning.

\textbf{A.\, Model-agnostic constrastive explanations.}
Explaining the decision making process of a complex ML model is crucial to many ML-assisted high-stakes applications, such as job hiring, financial loan application and judicial sentence. When generating local explanations for the prediction of an ML model on a specific data sample, one common practice is to leverage the information of its input gradient for sensitivity analysis of the model prediction. 
For ML models that do not have explicit functions for computing input gradients, such as access-limited APIs or rule-based systems, ZO optimization enables the generation of local explanations using model queries without the knowledge of the gradient information. Moreover, even when the input gradient can be obtained via ML platforms such as TensorFlow and PyTorch, the gradient computation is   platform-specific. In this case, ZO optimization has the advantage of alleviating platform dependency when developing multi-platform explanation methods, as it only depends on model inference results.

\begin{figure}[htb]
	\centering
	\begin{tabular}{c}
	   \hspace*{-0.15in}  \includegraphics[width=0.63\columnwidth]{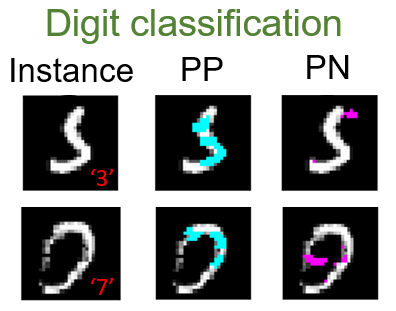} \vspace*{-0.015in} 
  \\
  (a)\\
		   \hspace*{-0.15in}      \includegraphics[width=1\columnwidth]{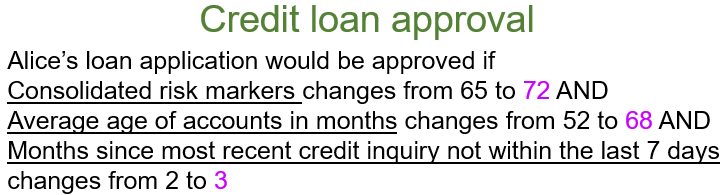} \vspace*{-0.05in} \\
		   	   
	     (b)
	\end{tabular}
		\caption{Constrastive explanations generated by ZO optimization methods. (a) For hand-written digit classification, the red digit class on the corner of an input sample shows the model prediction of the instance. The pixels highlighted by the cyan color are the pertinent positive (PP) supporting the original prediction.  The pixels highlighted by the purple color are the pertinent negative (PN) that will alter the model prediction when added to the original instance. (b) For the credit loan application, the PN of an applicant (Alice) is used to explain the necessary modifications on a subset of the original features in order to change the model prediction from `denial' to `approval'.
        }
  \label{Fig_CEM}
  \end{figure}


Here, we apply ZO optimization to generating contrastive explanations \cite{dhurandhar2018explanations} for two ML applications -- handwritten digit classification and loan approval. Contrastive explanations consist of two components derived from a given data sample for explaining the model prediction, i.e., a pertinent positive (PP) that is minimally and sufficiently present to keep the same prediction of the original input sample, and a pertinent negative (PN) that is minimally and necessarily absent to alter the model prediction.
The process of finding PP and PN is formulated as a sparsity-driven and data-perturbation based optimization problem guided by the model prediction outcomes \cite{dhurandhar2019model}, which can be solved by ZO optimization methods. Fig. \ref{Fig_CEM} shows the contrastive explanations generated from black-box neural network models by ZO-GD using the objective functions in \cite{dhurandhar2019model}. For hand-written digit classification, the PP identifies a subset of pixels such that their presence is minimally sufficient for   model prediction. Moreover, the PN identifies a subset of pixels such that their absence is minimally necessary for altering model prediction. The PP and PN  together constitute a contrastive explanation for interpreting model prediction. Similarly, for credit loan approval task trained on the FICO explainable machine learning challenge dataset\footnote{\url{https://community.fico.com/s/explainable-machine-learning-challenge}} based on a neural network model, the PN generated for an applicant (Alice) can be used to explain how the model would alter the recommendation from `denial' to `approval' based on Alice's loan application profile\footnote{Please refer to \url{https://aix360.mybluemix.net} for more details.}.

\textbf{B.\,Policy search in reinforcement learning.}
Reinforcement learning aims to determine given a state which action to take (or policy) in order to maximize the reward. 
One of the most popular policy search approaches is the model-free policy search, where agent learns parameterized policies from sampled trajectories without needing to learn the model of the underlying dynamics. Model-free policy search updates the parameters such that trajectories with higher reward are more likely to be obtained when following the updated policy~\cite{kober2013reinforcement}. Traditional policy search methods, such as REINFORCE~\cite{williams1992simple}, rely on randomized exploration in the action space to compute an estimated direction of improvement. These methods (referred to as policy gradient methods) then leverage the first order information of the policy (or Jacobian) to update its parameters to maximize the reward. Note that the chance of finding a sequence of actions resulting in high total reward decreases as the horizon length increases and thus policy gradient methods often exhibit high variance and result in  large sample complexity~\cite{zhao2011analysis}.

To alleviate these problems, ZO policy search methods, which directly optimize over policy parameter space, have emerged as an alternative to policy gradient. More specifically, ZO policy search methods seek to directly optimize the total reward in the space of parameters by employing   finite-difference  methods to compute estimates of the gradient with respect to policy parameters~\cite{kober2013reinforcement,rl1,rl7,pmlr-v89-malik19a,vemula2019contrasting}. These methods are fully zeroth-order, i.e., they do not  exploit   first-order information of the policy, the reward, or the dynamics. Interestingly, it has been observed that although policy gradient methods leverage more information, ZO policy search methods often perform better empirically.
{In particular, the work \cite{pmlr-v89-malik19a} characterized the convergence rate of ZO  policy optimization  when applied to linear-quadratic systems.}
And the work  \cite{vemula2019contrasting} theoretically showed that the complexity of exploration in the action space (using policy gradients) depends on both the dimensionality of the action space and the horizon length, as opposed to, the complexity of exploration in the parameter space (using ZO methods) depends only on the dimensionality of the parameter space. 

\textbf{C.\, Automated ML.}
The success of ML relies heavily on selecting the right pipeline algorithms for the problem at hand, and on setting its hyperparameters. Automated ML (AutoML) automates the process of model selection and hyperparameter optimization. It offers the advantages of producing simpler solutions, faster creation of those solutions, and models that often outperform hand-designed machine learning models. One could view AutoML as the process of   optimization of an unknown black-box function. Recently, several Bayesian optimization (BO) approaches have been proposed for AutoML~\cite{snoek2012practical,liu2019admm}. BO  works by building a probabilistic surrogate via Gaussian
process (GP)
for the objective function, and then using an acquisition function defined from this surrogate to decide where to sample.
However, BO suffers from a computational bottleneck: an internal {first-order}  solver is required to determine the  parameters of the GP model     by 
  maximizing the   log marginal likelihood of the current   function evaluations at each iteration of BO. The first-order solver is slow due to the difficulty of computing the gradient of the log-likelihood function with respect to the parameters of GP. To circumvent this difficulty, the ZO optimization algorithm can be used  to determine the hyperparameters and thus to accelerate  BO in AutoML  \cite{liu2019admm}. 
  {In the context of meta-learning, ZO optimization has also been leveraged to obviate the need for determining   computationally-intensive high-order derivatives during meta-training \cite{Song2020ES-MAML}. Lastly, we note that ZO optimization can be integrated with learning-to-optimize (L2O), which models the optimizer through a trainable DNN-based  meta-learner \cite{Ruan2020Learning,NIPS2019_9641}.}

\section{Open Questions and Discussions}
\label{sec: open_discussion}

Although there has been a great deal of progress on the design, theoretical analysis,   and applications of ZO optimization,  many  questions  and  challenges  still  remain. 

\textbf{A.\, ZO optimization with non-smooth objectives.}
 There exists a
  gap between the theoretical analysis of ZO optimizers and practical ML/DL applications with non-smooth objectives, where the former usually  requires the smoothness   of the objective function. There are {two} possible means of relaxing the smoothness assumption.
  First, the randomized smoothing technique
  ensures that  the convolution of two functions is at least as smooth as the
  smoothest of the two original functions. 
  Thus, $f_{\mu}$ is smooth even if $f$ is non-smooth in \eqref{eq: fmu}.  
 This motivates the technique of \textit{double randomization} that approximates a subgradient of a non-smooth objective function \cite{duchi2015optimal}, where
  an extra  randomized perturbation is introduced to prevent  drawing points from non-smooth regions of $f$. The downside of double randomization is the increase of function query complexity.
  Second, a model-based trust region method can be leveraged to  approximate the subgradient/gradient using linear or quadratic   interpolation   \cite{Larson_2019,berahas2019theoretical}.
 This leads to the  general approach of gradient estimation without imposing extra assumptions on the objective function. However, it increases the computation cost due to the need to solve nested regression problems.
  

%

\textbf{B.\, ZO optimization with black-box constraints.}
 The current   work on ZO optimization is   restricted to black-box objective functions with white-box  constraints. In the presence of black-box constraints, the introduction of barrier functions (instead of constraints) \cite{boyd2004convex} in the objective could be a potential solution.
 One could also employ the method of multipliers to reformulate black-box constraints as   regularization functions in the objective function \cite{liu2019admm}. 

\textbf{C.\, ZO optimization for  privacy-preserving  distributed learning.}
To protect the sensitive information of data   in the context of distributed learning,
it is common to add `noise' (randomness) into gradients of individual cost functions of  agents, known as  message-perturbing privacy strategy \cite{song2013stochastic}.
The level of privacy is often evaluated by  {differential privacy} (DP). A high degree of DP prevents
the adversary from gaining   meaningful personal information of
any individuals. Similarly,  ZO optimization also conceals the gradient information and allows the use of noisy gradient estimates that are constructed  from function values. Thus, one interesting  question is: can ZO optimization   be designed with privacy guarantees? In a more general sense,  it would be worthwhile to examine what roles ZO optimization plays   in the   privacy-preserving and Byzantine-tolerant federated learning setting.

\textbf{D.\, ZO optimization and automatic differentiation.}
Automatic differentiation (AD) provides a way for
efficiently and accurately evaluating derivatives of numeric functions, which are expressed as computer
programs \cite{baydin2018automatic}. The backpropagation algorithm, used for training neural networks, can be regarded as a specialized instance of AD under the reverse mode. AD decomposes the derivative of the complex function into   sub-derivatives of   constituent operations through
 the chain rule. When a sub-derivative is infeasible or difficult to compute, the ZO gradient estimation techniques could be integrated with AD. In particular, when the high-order derivatives (beyond gradient) are required, e.g., model-agnostic meta-learning \cite{finn2017model}, ZO optimization could help to overcome the derivative bottleneck.


\textbf{E.\, ZO optimization for discrete variables.} 
Many machine learning and signal processing tasks involve handling discrete variables, such as texts, graphs, sets and categorical data. In addition to the technique of relaxation to continuous values, it is worthwhile to explore and design ZO algorithms that directly operate on discrete domains.

\textbf{F.\, Tight convergence rates of ZO methods.}
Although the  optimal rate for ZO unconstrained convex optimization was studied in  \cite{duchi2015optimal}, there remain   many open questions on seeking the optimal rates, or associated tight lower bounds,   for general cases of ZO   constrained nonconvex optimization.

\section{Conclusions}
\label{sec: conclusion}
In this survey paper, we discussed various variants of ZO gradient estimators and focused on their statistical modelling as this leads to general ZO algorithms. We also provided an extensive comparison of ZO algorithms and discussed their iteration and function query complexities. 
Furthermore, we  presented numerous emerging applications of ZO optimization in signal processing and machine learning. 
Finally, we highlighted some unsolved research challenges in ZO optimization research and presented some promising future research directions.

\section*{Acknowledgement}
{\small B. Kailkhura was performed under the auspices of the U.S. Department of Energy by Lawrence Livermore National Laboratory under Contract DE-AC52-07NA27344.
}

{{
\bibliographystyle{IEEEbib}
\bibliography{Ref_ZO,refs2,Ref_Sijia,ref,ref_bhavya,IEEEabrv}

\begin{thebibliography}{10}

\bibitem{kiefer1952stochastic}
J.~Kiefer, J.~Wolfowitz, et~al.,
\newblock ``Stochastic estimation of the maximum of a regression function,''
\newblock {\em The Annals of Mathematical Statistics}, vol. 23, no. 3, pp.
  462--466, 1952.

\bibitem{spall1987stochastic}
J.~C. Spall,
\newblock ``A stochastic approximation technique for generating maximum
  likelihood parameter estimates,''
\newblock in {\em American Control Conference}, 1987, pp. 1161--1167.

\bibitem{nocedal2006numerical}
J.~Nocedal and S.~Wright,
\newblock {\em Numerical optimization},
\newblock Springer Science \& Business Media, 2006.

\bibitem{conn2009introduction}
A.~R. Conn, K.~Scheinberg, and L.~N. Vicente,
\newblock {\em Introduction to derivative-free optimization}, vol.~8,
\newblock SIAM, 2009.

\bibitem{rios2013derivative}
L.~M. Rios and N.~V. Sahinidis,
\newblock ``Derivative-free optimization: a review of algorithms and comparison
  of software implementations,''
\newblock {\em Journal of Global Optimization}, vol. 56, no. 3, pp. 1247--1293,
  2013.

\bibitem{Larson_2019}
J.~Larson, M.~Menickelly, and S.~M. Wild,
\newblock ``Derivative-free optimization methods,''
\newblock {\em Acta Numerica}, vol. 28, pp. 287–404, May 2019.

\bibitem{nelder1965simplex}
J.~A. Nelder and R.~Mead,
\newblock ``A simplex method for function minimization,''
\newblock {\em The Computer Journal}, vol. 7, no. 4, pp. 308--313, 1965.

\bibitem{fermi1942}
E.~Fermi and N.~Metropolis,
\newblock ``Numerical solution of a minimum problem,''
\newblock {\em Technical Report LA-1492,
  \url{https://hdl.handle.net/2027/mdp.39015086487645}}, 1952.

\bibitem{torczon1991convergence}
V.~Torczon,
\newblock ``On the convergence of the multidirectional search algorithm,''
\newblock {\em SIAM Journal on Optimization}, vol. 1, no. 1, pp. 123--145,
  1991.

\bibitem{bortz1998simplex}
D.~M. Bortz and C.~T. Kelley,
\newblock ``The simplex gradient and noisy optimization problems,''
\newblock in {\em Computational Methods for Optimal Design and Control}, pp.
  77--90. Springer, 1998.

\bibitem{conn2000trust}
A.~R. Conn, N.~I. Gould, and P.~L. Toint,
\newblock {\em Trust region methods}, vol.~1,
\newblock {SIAM}, 2000.

\bibitem{vaz2009pswarm}
A.~I.~F. Vaz and L.~N. Vicente,
\newblock ``{PS}warm: a hybrid solver for linearly constrained global
  derivative-free optimization,''
\newblock {\em Optimization Methods \& Software}, vol. 24, no. 4-5, pp.
  669--685, 2009.

\bibitem{goldberg1988genetic}
D.~E. Goldberg and J.~H. Holland,
\newblock ``Genetic algorithms and machine learning,''
\newblock {\em Machine Learning}, vol. 3, no. 2-3, pp. 95--99, 1988.

\bibitem{shahriari2016taking}
B.~Shahriari, K.~Swersky, Z.~Wang, R.~P. Adams, and N.~De~Freitas,
\newblock ``Taking the human out of the loop: A review of bayesian
  optimization,''
\newblock {\em Proceedings of the IEEE}, vol. 104, no. 1, pp. 148--175, 2016.

\bibitem{powell1994direct}
M.~J.~D. Powell,
\newblock ``A direct search optimization method that models the objective and
  constraint functions by linear interpolation,''
\newblock in {\em Advances in optimization and numerical analysis}, pp. 51--67.
  Springer, 1994.

\bibitem{scipy2001}
E.~Jones, T.~Oliphant, P.~Peterson, et~al.,
\newblock ``{SciPy}: Open source scientific tools for {Python},'' 2001.

\bibitem{deng2009imagenet}
J.~Deng, W.~Dong, R.~Socher, L.-J. Li, K.~Li, and F.-F. Li,
\newblock ``Imagenet: A large-scale hierarchical image database,''
\newblock in {\em IEEE CVPR}, 2009, pp. 248--255.

\bibitem{ghadimi2013stochastic}
S.~Ghadimi and G.~Lan,
\newblock ``Stochastic first-and zeroth-order methods for nonconvex stochastic
  programming,''
\newblock {\em SIAM Journal on Optimization}, vol. 23, no. 4, pp. 2341--2368,
  2013.

\bibitem{nesterov2015random}
Y.~Nesterov and V.~Spokoiny,
\newblock ``Random gradient-free minimization of convex functions,''
\newblock {\em Foundations of Computational Mathematics}, vol. 2, no. 17, pp.
  527--566, 2015.

\bibitem{flaxman2005online}
A.~D. Flaxman, A.~T. Kalai, and H.~B. McMahan,
\newblock ``Online convex optimization in the bandit setting: {G}radient
  descent without a gradient,''
\newblock in {\em Proc. 16th Annual ACM-SIAM Symposium on Discrete algorithms},
  2005, pp. 385--394.

\bibitem{duchi2015optimal}
J.~C. Duchi, M.~I. Jordan, M.~J. Wainwright, and A.~Wibisono,
\newblock ``Optimal rates for zero-order convex optimization: The power of two
  function evaluations,''
\newblock {\em {IEEE} Trans. Inf. Theory}, vol. 61, no. 5, pp. 2788--2806,
  2015.

\bibitem{goodfellow2015explaining}
I.~Goodfellow, J.~Shlens, and C.~Szegedy,
\newblock ``Explaining and harnessing adversarial examples,''
\newblock {\em ICLR}, 2015.

\bibitem{chen2017zoo}
P.-Y. Chen, H.~Zhang, Y.~Sharma, J.~Yi, and C.-J. Hsieh,
\newblock ``Zoo: Zeroth order optimization based black-box attacks to deep
  neural networks without training substitute models,''
\newblock in {\em Proc. 10th ACM Workshop on Artificial Intelligence and
  Security}, 2017, pp. 15--26.

\bibitem{ilyas2018blackbox}
A.~Ilyas, K.~Engstrom, A.~Athalye, and J.~Lin,
\newblock ``Black-box adversarial attacks with limited queries and
  information,''
\newblock in {\em ICML}, July 2018.

\bibitem{dhurandhar2019model}
A.~Dhurandhar, T.~Pedapati, A.~Balakrishnan, et~al.,
\newblock ``Model agnostic contrastive explanations for structured data,''
\newblock {\em arXiv preprint arXiv:1906.00117}, 2019.

\bibitem{liu2019admm}
S.~Liu, P.~Ram, D.~Vijaykeerthy, et~al.,
\newblock ``An {ADMM} based framework for {AutoML} pipeline configuration,''
\newblock {\em AAAI}, 2019.

\bibitem{liu2017zeroth}
S.~Liu, J.~Chen, P.-Y. Chen, and A.~O. Hero,
\newblock ``Zeroth-order online {ADMM}: Convergence analysis and
  applications,''
\newblock in {\em AISTATS}, 2018, vol.~84, pp. 288--297.

\bibitem{moosavi2019robustness}
S.-M. Moosavi-Dezfooli, A.~Fawzi, J.~Uesato, and P.~Frossard,
\newblock ``Robustness via curvature regularization, and vice versa,''
\newblock in {\em IEEE CVPR}, 2019, pp. 9078--9086.

\bibitem{Song2020ES-MAML}
X.~Song, W.~Gao, Y.~Yang, K.~Choromanski, A.~Pacchiano, and Y.~Tang,
\newblock ``Es-maml: Simple hessian-free meta learning,''
\newblock in {\em ICLR}, 2020.

\bibitem{Tsai_BAR}
Y.-Y. Tsai, P.-Y. Chen, and T.-Y. Ho,
\newblock ``Transfer learning without knowing: Reprogramming black-box machine
  learning models with scarce data and limited resources,''
\newblock {\em ICML}, 2020.

\bibitem{berahas2019theoretical}
A.~S. Berahas, L.~Cao, K.~Choromanski, and K.~Scheinberg,
\newblock ``A theoretical and empirical comparison of gradient approximations
  in derivative-free optimization,''
\newblock {\em arXiv preprint arXiv:1905.01332}, 2019.

\bibitem{liu2018_NIPS}
S.~Liu, B.~Kailkhura, P.-Y. Chen, P.~Ting, S.~Chang, and L.~Amini,
\newblock ``Zeroth-order stochastic variance reduction for nonconvex
  optimization,''
\newblock {\em NeurIPS}, 2018.

\bibitem{spall1992multivariate}
J.~C. Spall et~al.,
\newblock ``Multivariate stochastic approximation using a simultaneous
  perturbation gradient approximation,''
\newblock {\em IEEE transactions on automatic control}, vol. 37, no. 3, pp.
  332--341, 1992.

\bibitem{lian2016comprehensive}
X.~Lian, H.~Zhang, C.-J. Hsieh, Y.~Huang, and J.~Liu,
\newblock ``A comprehensive linear speedup analysis for asynchronous stochastic
  parallel optimization from zeroth-order to first-order,''
\newblock in {\em NeurIPS}, 2016, pp. 3054--3062.

\bibitem{shamir2017optimal}
O.~Shamir,
\newblock ``An optimal algorithm for bandit and zero-order convex optimization
  with two-point feedback,''
\newblock {\em JMLR}, vol. 18, no. 52, pp. 1--11, 2017.

\bibitem{liu2018signsgd}
S.~Liu, P.-Y. Chen, X.~Chen, and M.~Hong,
\newblock ``sign{SGD} via zeroth-order oracle,''
\newblock in {\em ICLR}, 2019.

\bibitem{gu2016zeroth}
B.~Gu, Z.~Huo, and H.~Huang,
\newblock ``Zeroth-order asynchronous doubly stochastic algorithm with variance
  reduction,''
\newblock {\em arXiv preprint arXiv:1612.01425}, 2016.

\bibitem{liu2018stochastic}
L.~Liu, M.~Cheng, C.-J. Hsieh, and D.~Tao,
\newblock ``Stochastic zeroth-order optimization via variance reduction
  method,''
\newblock {\em arXiv preprint arXiv:1805.11811}, 2018.

\bibitem{JWL19}
K.~Ji, Z.~Wang, Y.~Zhou, and Y.~Liang,
\newblock ``Improved zeroth-order variance reduced algorithms and analysis for
  nonconvex optimization,''
\newblock in {\em ICML}, 2019, vol.~97, pp. 3100--3109.

\bibitem{ye2018hessian}
H.~Ye, Z.~Huang, C.~Fang, C.~J. Li, and T.~Zhang,
\newblock ``Hessian-aware zeroth-order optimization for black-box adversarial
  attack,''
\newblock {\em arXiv preprint arXiv:1812.11377}, 2018.

\bibitem{balasubramanian2019zeroth}
K.~Balasubramanian and S.~Ghadimi,
\newblock ``Zeroth-order nonconvex stochastic optimization: Handling
  constraints, high-dimensionality, and saddle-points,''
\newblock {\em arXiv preprint arXiv:1809.06474}, pp. 651--676, 2019.

\bibitem{cheng2019sign}
M.~Cheng, S.~Singh, P.-Y. Chen, S.~Liu, and C.-J. Hsieh,
\newblock ``Sign-{OPT}: A query-efficient hard-label adversarial attack,''
\newblock {\em ICLR}, 2020.

\bibitem{ghadimi2016mini}
S.~Ghadimi, G.~Lan, and H.~Zhang,
\newblock ``Mini-batch stochastic approximation methods for nonconvex
  stochastic composite optimization,''
\newblock {\em Mathematical Programming}, vol. 155, no. 1-2, pp. 267--305,
  2016.

\bibitem{balasubramanian2018zeroth}
K.~Balasubramanian and S.~Ghadimi,
\newblock ``Zeroth-order (non)-convex stochastic optimization via conditional
  gradient and gradient updates,''
\newblock in {\em NeurIPS}, 2018, pp. 3455--3464.

\bibitem{chen2019zo}
X.~Chen, S.~Liu, K.~Xu, X.~Li, X.~Lin, M.~Hong, and D.~Cox,
\newblock ``{ZO-AdaMM}: Zeroth-order adaptive momentum method for black-box
  optimization,''
\newblock in {\em NeurIPS}, 2019, pp. 7202--7213.

\bibitem{chen2018frank}
J.~Chen, J.~Yi, and Q.~Gu,
\newblock ``A {F}rank-{W}olfe framework for efficient and effective adversarial
  attacks,''
\newblock in {\em AAAI}, 2020.

\bibitem{sahu2018towards}
A.~K. Sahu, M.~Zaheer, and S.~Kar,
\newblock ``Towards gradient free and projection free stochastic
  optimization,''
\newblock {\em AISTATS}, 2019.

\bibitem{chen2018convergence}
X.~Chen, S.~Liu, R.~Sun, and M.~Hong,
\newblock ``On the convergence of a class of adam-type algorithms for
  non-convex optimization,''
\newblock in {\em ICLR}, 2019.

\bibitem{Zhao_2019_ICCV}
P.~Zhao, S.~Liu, P.-Y. Chen, N.~Hoang, K.~Xu, B.~Kailkhura, and X.~Lin,
\newblock ``On the design of black-box adversarial examples by leveraging
  gradient-free optimization and operator splitting method,''
\newblock in {\em ICCV}, 2019.

\bibitem{gao2014information}
X.~Gao, B.~Jiang, and S.~Zhang,
\newblock ``On the information-adaptive variants of the {ADMM}: an iteration
  complexity perspective,''
\newblock {\em Optimization Online}, vol. 12, 2014.

\bibitem{huang2019zeroth}
F.~Huang, S.~Gao, S.~Chen, and H.~Huang,
\newblock ``Zeroth-order stochastic alternating direction method of multipliers
  for nonconvex nonsmooth optimization,''
\newblock {\em IJCAI}, 2019.

\bibitem{liu2019min}
S.~Liu, S.~Lu, X.~Chen, Y.~Feng, K.~Xu, A.~Al-Dujaili, M.~Hong, and U.-M.
  O'Reilly,
\newblock ``Min-max optimization without gradients: Convergence and
  applications to adversarial {ML},''
\newblock in {\em ICML}, 2020.

\bibitem{wang2020zeroth}
Z.~Wang, K.~Balasubramanian, S.~Ma, and M.~Razaviyayn,
\newblock ``Zeroth-order algorithms for nonconvex minimax problems with
  improved complexities,''
\newblock {\em arXiv preprint arXiv:2001.07819}, 2020.

\bibitem{sahu2018distributed}
A.~K. Sahu, D.~Jakovetic, D.~Bajovic, and S.~Kar,
\newblock ``Distributed zeroth order optimization over random networks: A
  kiefer-wolfowitz stochastic approximation approach,''
\newblock in {\em IEEE CDC}, 2018, pp. 4951--4958.

\bibitem{yuan2015zeroth}
D.~Yuan, D.~W.~C. Ho, and S.~Xu,
\newblock ``Zeroth-order method for distributed optimization with approximate
  projections,''
\newblock {\em {IEEE} Trans. Neural Netw.}, vol. 27, no. 2, pp. 284--294, 2015.

\bibitem{yu2019distributed}
Z.~Yu, D.~W.~C. Ho, and D.~Yuan,
\newblock ``Distributed randomized gradient-free mirror descent algorithm for
  constrained optimization,''
\newblock {\em arXiv preprint arXiv:1903.04157}, 2019.

\bibitem{hajinezhad2017zeroth}
D.~{Hajinezhad}, M.~{Hong}, and A.~{Garcia},
\newblock ``Zone: Zeroth-order nonconvex multiagent optimization over
  networks,''
\newblock {\em {IEEE} Trans. Autom. Control}, vol. 64, no. 10, pp. 3995--4010,
  2019.

\bibitem{tang2019distributed}
Y.~Tang and N.~Li,
\newblock ``Distributed zero-order algorithms for nonconvex multi-agent
  optimization,''
\newblock in {\em Allerton}, 2019, pp. 781--786.

\bibitem{wangdu18}
Y.~Wang, S.~Du, S.~Balakrishnan, and A.~Singh,
\newblock ``Stochastic zeroth-order optimization in high dimensions,''
\newblock in {\em AISTATS}. April 2018, vol.~84, pp. 1356--1365, PMLR.

\bibitem{golovin2019gradientless}
D.~Golovin, J.~Karro, G.~Kochanski, C.~Lee, X.~Song, and Q.~Zhang,
\newblock ``Gradientless descent: High-dimensional zeroth-order optimization,''
\newblock in {\em ICLR}, 2020.

\bibitem{li2020zeroth}
J.~Li, K.~Balasubramanian, and S.~Ma,
\newblock ``Zeroth-order optimization on {R}iemannian manifolds,''
\newblock {\em arXiv preprint arXiv:2003.11238}, 2020.

\bibitem{hazan2016introduction}
E.~Hazan,
\newblock ``Introduction to online convex optimization,''
\newblock {\em Foundations and Trends{\textregistered} in Optimization}, vol.
  2, no. 3-4, pp. 157--325, 2016.

\bibitem{vlatakis2019efficiently}
E.-V. Vlatakis-Gkaragkounis, L.~Flokas, and G.~Piliouras,
\newblock ``Efficiently avoiding saddle points with zero order methods: No
  gradients required,''
\newblock in {\em NeurIPS}, 2019, pp. 10066--10077.

\bibitem{reddi2016proximal}
S.~J. Reddi, S.~Sra, B.~Poczos, and A.~J. Smola,
\newblock ``Proximal stochastic methods for nonsmooth nonconvex finite-sum
  optimization,''
\newblock in {\em NeurIPS}, 2016, pp. 1145--1153.

\bibitem{carlini2017towards}
N.~Carlini and D.~Wagner,
\newblock ``Towards evaluating the robustness of neural networks,''
\newblock in {\em IEEE Symposium on Security and Privacy}, 2017, pp. 39--57.

\bibitem{tu2018autozoom}
C.-C. Tu, P.~Ting, P.-Y. Chen, S.~Liu, H.~Zhang, J.~Yi, C.-J. Hsieh, and S.-M.
  Cheng,
\newblock ``Autozoom: Autoencoder-based zeroth order optimization method for
  attacking black-box neural networks,''
\newblock in {\em AAAI}, 2019.

\bibitem{brendel2017decision}
W.~Brendel, J.~Rauber, and M.~Bethge,
\newblock ``Decision-based adversarial attacks: Reliable attacks against
  black-box machine learning models,''
\newblock {\em ICLR}, 2018.

\bibitem{cheng2018query}
M.~Cheng, T.~Le, P.-Y. Chen, J.~Yi, H.~Zhang, and C.-J. Hsieh,
\newblock ``Query-efficient hard-label black-box attack: An optimization-based
  approach,''
\newblock {\em ICLR}, 2019.

\bibitem{zhao2020towards}
P.~Zhao, P.-Y. Chen, S.~Wang, and X.~Lin,
\newblock ``Towards query-efficient black-box adversary with zeroth-order
  natural gradient descent,''
\newblock {\em AAAI}, 2020.

\bibitem{xu2018structured}
K.~Xu, S.~Liu, P.~Zhao, et~al.,
\newblock ``Structured adversarial attack: Towards general implementation and
  better interpretability,''
\newblock in {\em ICLR}, 2019.

\bibitem{szegedy2016rethinking}
C.~Szegedy, V.~Vanhoucke, S.~Ioffe, J.~Shlens, and Z.~Wojna,
\newblock ``Rethinking the inception architecture for computer vision,''
\newblock in {\em IEEE CVPR}, 2016, pp. 2818--2826.

\bibitem{chen2017bandit}
T.~Chen and G.~B. Giannakis,
\newblock ``Bandit convex optimization for scalable and dynamic {IoT}
  management,''
\newblock {\em IEEE Internet of Things Journal}, 2018.

\bibitem{hero2011sensor}
A.~O. Hero and D.~Cochran,
\newblock ``Sensor management: Past, present, and future,''
\newblock {\em {IEEE} Sensors J.}, vol. 11, no. 12, pp. 3064--3075, 2011.

\bibitem{joshi2009sensor}
S.~Joshi and S.~Boyd,
\newblock ``Sensor selection via convex optimization,''
\newblock {\em {IEEE} Trans. Signal Process.}, vol. 57, no. 2, pp. 451--462,
  2009.

\bibitem{suzuki2013dual}
T.~Suzuki,
\newblock ``Dual averaging and proximal gradient descent for online alternating
  direction multiplier method,''
\newblock in {\em ICML}, 2013, pp. 392--400.

\bibitem{dhurandhar2018explanations}
A.~Dhurandhar, P.-Y. Chen, et~al.,
\newblock ``Explanations based on the missing: Towards contrastive explanations
  with pertinent negatives,''
\newblock in {\em NeurIPS}, 2018, pp. 592--603.

\bibitem{kober2013reinforcement}
J.~Kober, J.~A. Bagnell, and J.~Peters,
\newblock ``Reinforcement learning in robotics: A survey,''
\newblock {\em The International Journal of Robotics Research}, vol. 32, no.
  11, pp. 1238--1274, 2013.

\bibitem{williams1992simple}
R.~J Williams,
\newblock ``Simple statistical gradient-following algorithms for connectionist
  reinforcement learning,''
\newblock {\em Machine learning}, vol. 8, no. 3-4, pp. 229--256, 1992.

\bibitem{zhao2011analysis}
T.~Zhao, H.~Hachiya, G.~Niu, and M.~Sugiyama,
\newblock ``Analysis and improvement of policy gradient estimation,''
\newblock in {\em NeurIPS}, 2011, pp. 262--270.

\bibitem{rl1}
T.~Salimans, H.~Ho, X.~Chen, S.~Sidor, and I.~Sutskever,
\newblock ``Evolution strategies as a scalable alternative to reinforcement
  learning,''
\newblock {\em arXiv preprint arXiv:1703.03864}, 2017.

\bibitem{rl7}
F.~Sehnke, C.~Osendorfer, Thomas R{\"u}ckstie{\ss}, A.~Graves, J.~Peters, and
  J.~Schmidhuber,
\newblock ``Parameter-exploring policy gradients,''
\newblock {\em Neural Networks}, vol. 23, no. 4, pp. 551--559, 2010.

\bibitem{pmlr-v89-malik19a}
D.~Malik, A.~Pananjady, K.~Bhatia, K.~Khamaru, P.~Bartlett, and M.~Wainwright,
\newblock ``Derivative-free methods for policy optimization: Guarantees for
  linear quadratic systems,''
\newblock in {\em Proceedings of Machine Learning Research}, 16--18 Apr 2019,
  vol.~89, pp. 2916--2925.

\bibitem{vemula2019contrasting}
A.~Vemula, W.~Sun, and J.~A. Bagnell,
\newblock ``Contrasting exploration in parameter and action space: A
  zeroth-order optimization perspective,''
\newblock in {\em Proceedings of Machine Learning Research}, 16--18 Apr 2019,
  vol.~89, pp. 2926--2935.

\bibitem{snoek2012practical}
J.~Snoek, H.~Larochelle, and R.~P. Adams,
\newblock ``Practical bayesian optimization of machine learning algorithms,''
\newblock in {\em NeurIPS}, 2012, pp. 2951--2959.

\bibitem{Ruan2020Learning}
Y.~Ruan, Y.~Xiong, S.~Reddi, S.~Kumar, and C.-J. Hsieh,
\newblock ``Learning to learn by zeroth-order oracle,''
\newblock in {\em ICLR}, 2020.

\bibitem{NIPS2019_9641}
Y.~Cao, T.~Chen, Z.~Wang, and Y.~Shen,
\newblock ``Learning to optimize in swarms,''
\newblock in {\em NeurIPS}, 2019, pp. 15044--15054.

\bibitem{boyd2004convex}
S.~Boyd and L.~Vandenberghe,
\newblock {\em Convex optimization},
\newblock Cambridge university press, 2004.

\bibitem{song2013stochastic}
S.~Song, K.~Chaudhuri, and A.~D. Sarwate,
\newblock ``Stochastic gradient descent with differentially private updates,''
\newblock in {\em IEEE GlobalSIP}, 2013, pp. 245--248.

\bibitem{baydin2018automatic}
A.~G. Baydin, B.~A. Pearlmutter, A.~A. Radul, and J.~M. Siskind,
\newblock ``Automatic differentiation in machine learning: A survey,''
\newblock {\em JMLR}, vol. 18, no. 1, pp. 5595--5637, Jan. 2017.

\bibitem{finn2017model}
C.~Finn, P.~Abbeel, and S.~Levine,
\newblock ``Model-agnostic meta-learning for fast adaptation of deep
  networks,''
\newblock in {\em ICML}, 2017, pp. 1126--1135.

\end{thebibliography}
}}

\ifCLASSOPTIONcaptionsoff
  \newpage
\fi

\end{document}